\documentclass{article}

\PassOptionsToPackage{numbers, compress}{natbib}

\usepackage[preprint]{neurips_2022}

\usepackage[utf8]{inputenc} 
\usepackage[T1]{fontenc}    
\usepackage{hyperref}       
\usepackage{url}            
\usepackage{booktabs}       
\usepackage{amsfonts}       
\usepackage{nicefrac}       
\usepackage{microtype}      
\usepackage[usenames,dvipsnames]{color}
\usepackage{xcolor}         
\usepackage{multicol}
\usepackage{wrapfig}
\usepackage{bbm}
\usepackage{algorithm} 
\usepackage{algpseudocode} 
\usepackage{amsmath} 
\usepackage{graphicx}
\usepackage{booktabs}
\usepackage{subcaption}
\usepackage{fancyvrb}
\usepackage{listings}
\usepackage[normalem]{ulem}

\renewcommand{\citet}[1]{\citep{#1}}
\renewcommand{\citeauthor}[1]{\citep{#1}}

\newcommand\rat{\text{rat}}
\definecolor{darkspringgreen}{rgb}{0.09, 0.45, 0.27}

\title{STaR: Self-Taught Reasoner \\ {
\Large
Bootstrapping Reasoning With Reasoning
}}

\author{%
  Eric Zelikman$^{*1}$, 
  Yuhuai Wu$^{*12}$, 
  Jesse Mu$^1$, 
  Noah D. Goodman$^1$
    \\
  $^1$Department of Computer Science, Stanford University\\
  $^2$ Google Research \\
  \texttt{\{ezelikman, yuhuai, muj, ngoodman\}@stanford.edu} \\
}

\begin{document}
\let\oldparagraph\paragraph
\renewcommand{\paragraph}{\vspace{-0.75em}\oldparagraph}

\def\thefootnote{*}\footnotetext{These authors contributed equally to this work}\def\thefootnote{\arabic{footnote}}
\maketitle
\begin{abstract}
Generating step-by-step "chain-of-thought" rationales improves language model performance on complex reasoning tasks like mathematics or commonsense question-answering. However, inducing language model rationale generation currently requires either constructing massive rationale datasets or sacrificing accuracy by using only few-shot inference.
We propose a technique to iteratively leverage a small number of rationale examples and a large dataset without rationales, to bootstrap the ability to perform successively more complex reasoning. 
This technique, the "Self-Taught Reasoner" (STaR), relies on a simple loop: generate rationales to answer many questions, prompted with a few rationale examples; if the generated answers are wrong, 
try again to generate a rationale given the correct answer; fine-tune on all the rationales that ultimately yielded correct answers; repeat.
We show that STaR significantly improves performance on multiple datasets compared to a model fine-tuned to directly predict final answers, and performs comparably to fine-tuning a 30$\times$ larger state-of-the-art language model on CommensenseQA. Thus, STaR lets a model improve itself by learning from its own generated reasoning.
\end{abstract}
\vspace{-9px}
\section{Introduction}
Human decision-making is often the result of extended chains of thought \citep{james1890principles,ericsson1984protocol}.
Recent work has shown that explicit intermediate reasoning (``rationales'') can improve large language model~(LLM) performance as well \citep{rajani2019explain, shwartz2020unsupervised, nye2021show, wei2022chain, marasovic2021few, lampinen2022can}. 
For example, \citet{nye2021show} demonstrated that LLMs explicitly trained to use ``scratchpads'' for intermediate steps can attain perfect in-distribution performance on arithmetic, and strong out-of-distribution generalization, while models trained to predict answers directly fail to do either. 
These works suggest that generating explicit rationales before giving a final answer (``rationale generation'') is valuable for LLMs across 
diverse
tasks including mathematical reasoning, commonsense reasoning, code evaluation, social bias inference, and natural language inference.
However, the two primary methods for inducing rationale generation both have serious drawbacks.

One approach to rationale generation is the construction of a fine-tuning dataset of rationales, either manually by human annotators or automatically with hand-crafted templates \citep{rajani2019explain,shwartz2020unsupervised,nye2021show,cobbe2021training}. Manual methods are expensive, and it is infeasible to construct such a dataset for each interesting problem \citep{rajani2019explain}. Meanwhile, template-based methods rely on automatically-generated rationales but only work when a general solution is already known \citep{nye2021show} or reasonable hard-coded heuristics can be made \citep{shwartz2020unsupervised}. 

An alternative is to leverage in-context learning by including only a few rationale examples in the language model prompt. This has been shown to improve accuracy on mathematical and symbolic reasoning tasks relative to prompting without rationales (``direct'' prompting) \citep{nye2021show,wei2022chain}. Yet, while few-shot techniques with rationales tend to outperform their non-reasoning counterparts, they generally substantially underperform models fine-tuned to directly predict answers using larger datasets \citep{nye2021show,wei2022chain}. 

\begin{figure}
\centering
\hspace{-20px}
\begin{subfigure}{.75\textwidth}
  \centering
    {
    \small
\hspace{3px}\includegraphics[width=\textwidth]{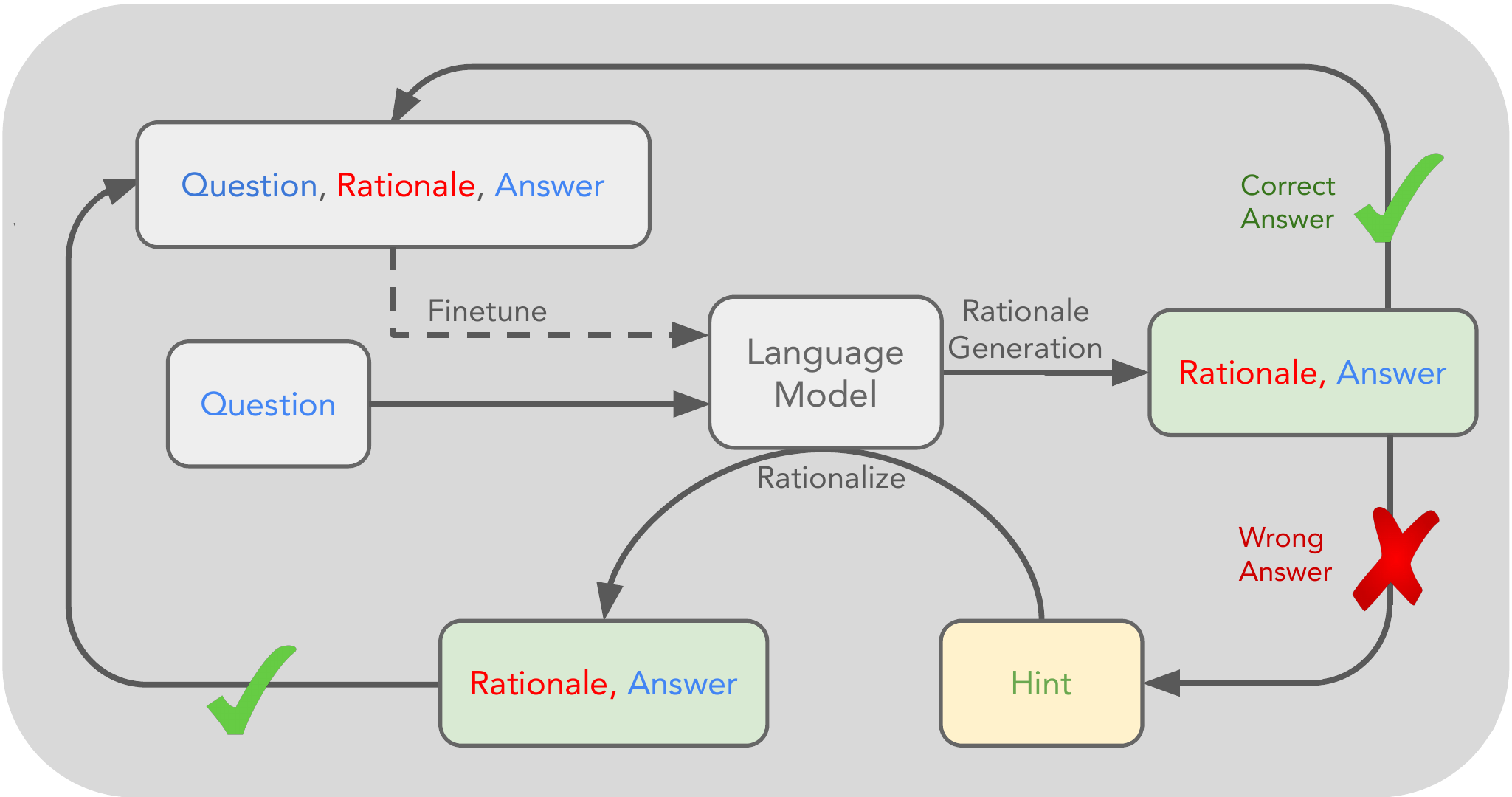}    }
\end{subfigure}
\hspace{5px}
\begin{subfigure}{.22\textwidth}
  \centering
    {
    \scriptsize
    \begin{Verbatim}[commandchars=+\[\]]
+makeblue[Q: What can be used]
+makeblue[to carry a small dog?]
+makeblue[Answer Choices:]
+makeblue[(a) swimming pool]
+makeblue[(b) basket]
+makeblue[(c) dog show]
+makeblue[(d) backyard]
+makeblue[(e) own home]
+makeblue[A:] +makered[The answer must be]
+makered[something that can be]
+makered[used to carry a small]
+makered[dog. Baskets are]
+makered[designed to hold things.]
+makered[Therefore, the answer]
+makered[is basket] +makeblue[(b)].
\end{Verbatim}
}
\end{subfigure}
\hspace{-20px}
\caption{An overview of STaR and a STaR-generated rationale on CommonsenseQA. We indicate the fine-tuning outer loop with a dashed line. 
The {\color{blue} questions} and ground truth {\color{blue} answers} are expected to be present in the dataset, while the {\color{Maroon} rationales} are generated using STaR.} 
\label{fig:overview}
\vspace{-15px}
\end{figure}

In this paper, we adopt a different approach: by leveraging the LLM's pre-existing reasoning ability, we iteratively \emph{bootstrap} the ability to generate high-quality rationales.
Specifically, we few-shot prompt a large language model to self-generate rationales and refine the model's ability further by fine-tuning on those rationales that lead to correct answers. We repeat this procedure, using the improved model to generate the next training set each time. 
This is a synergistic process, where improvements in rationale generation improve the training data, and improvements in training data further improve rationale generation. 

However, we find this loop eventually fails to solve any new problems in the training set because it receives no direct training signal for problems it fails to solve. 
To overcome this issue, we propose \textbf{rationalization}: for each problem that the model fails to answer correctly, we generate a new rationale by providing the model with the correct answer.
This lets the model reason backward---given the correct answer, the model can more easily generate a useful rationale. These rationales are then collected as part of the training data, which often improves overall accuracy.

We thus develop the Self-Taught Reasoner (STaR, Fig.~\ref{fig:overview}) method, a scalable bootstrapping method allowing models to learn to generate their own rationales, while also learning to solve increasingly difficult problems. In our method, we repeat the following process: in each iteration, first construct a finetuning dataset by attempting to solve the dataset using the current model's \textbf{rationale generation} ability; then, augment this dataset using \textbf{rationalization}, justifying ground-truth answers to problems the model failed to solve; finally, finetune the large language model on the combined dataset. 

Applying STaR on arithmetic, math word problems, and commonsense reasoning, we observe it is able to effectively translate a small number of few-shot prompts into a large rationale dataset,  yielding dramatic performance improvements. On CommonsenseQA \citep{talmor2019commonsenseqa}, we find STaR improves over both a few-shot baseline 
(+35.9\%)
and a baseline fine-tuned to directly predict answers 
(+12.5\%)
, and performs comparably to a fine-tuned model that is 30$\times$ larger (72.5\% vs.\ 73.0\%).

Thus, we make the following contributions:
\vspace{-5px}
\begin{enumerate}
    \itemsep0em 
    \item We propose a bootstrapping mechanism to iteratively generate a rationale dataset from a few initial examples with rationales---without needing to check new rationales' correctness.
    \item We complement \textbf{rationale generation} with \textbf{rationalization}, where a model is tasked with justifying an answer and then fine-tuned as if it had come up with the rationale without any hint. We show rationalization  accelerates and improves the bootstrapping process.
    \item We evaluate these techniques with a variety of ablations in both mathematical and commonsense reasoning domains.
    \item We propose what is, to our knowledge, the first technique to allow a  pre-trained large language model to iteratively use its language modeling capacity to improve itself.
\end{enumerate}
\section{Background and Related Work}

\paragraph{In-context Learning}
Recently, a collection of works has emerged exploring the capacity for large language models to perform in-context learning~\citep{brown2020language,wei2021finetuned}. In essence, in-context learning treats few-shot learning as a language modeling problem, by showing a few examples in the context (i.e.\ prompt), and allowing the model to learn and identify the pattern to apply to new examples. Some have studied in-context learning based on the language modeling objective in terms of Bayesian inference \citet{xie2021explanation} while others have attempted to describe the process more mechanistically in terms of ``induction heads'' \citep{olsson_elhage}. Moreover, differences in prompt configurations have been known to have dramatic effects on few-shot performance. Some have even found that replacing few-shot prompts with a ``soft prompt'' which can be optimized in embedding space results in noticeable gains \citep{lester2021power}. Instead of emphasizing the representation of the question, we focus on the model output; in particular, we focus on the model's ability to reason through a problem before coming to a conclusion.
\paragraph{Rationales}
One of the initial works on the impact of rationales on language model performance was \citet{rajani2019explain}, showing that training a language model on a dataset with explicit rationales preceding the answer could improve a model's ability to generate the final answer. However, this required many thousands of training examples to be manually annotated with human reasoning. Recently, \citet{nye2021show} demonstrated that step-by-step ``scratchpads'' can improve fine-tuned LLM performance and generalization on tasks such as arithmetic, polynomial evaluation, and program evaluation. Similarly, \citet{wei2022chain} used a single few-shot ``chain-of-thought'' reasoning prompt in order to improve model performance on a collection of tasks, without fine-tuning. Finally, \citet{polu2022formal} showed that a curriculum learning approach could help solve formal math problems, as long as 1) they were translated into Lean (a theorem-proving language \citep{moura2015lean}), 2) one could directly evaluate the validity of the proofs, 3) one could sample numerous potential solutions for each problem, 4) had trained a separate value function model, and 5) started with GPT-f (a model already fine-tuned on a large math dataset \citep{polu2020generative}). We note that there are many domains where these conditions do not all apply. In addition, works have aimed to explain why rationales have this beneficial effect: some have analyzed their impact from the perspective of latent variable models \citep{zhou2020towards} while others have provided formal proofs of the benefit of intermediate task supervision \citep{wies2022sub}. 

\paragraph{Iterated Learning} \looseness=-1
A variety of iterated learning algorithms have been proposed, where solutions or successful methods which are found are in turn used to find additional solutions \citep{anthony2017thinking, vani2021iterated, polu2022formal}. \citet{anthony2017thinking} introduced Expert Iteration (ExIt), a reinforcement learning technique serving as an inspiration for our approach. Essentially, it consists of a loop of self-play by an ``apprentice,'' followed by imitation learning with feedback from a slower ``expert'' and then the replacement of the expert with the now-improved apprentice. \citet{polu2022formal} builds off of ExIt for formal reasoning, while \citet{vani2021iterated} applies iterated learning to visual question answering using modular networks which can be combined compositionally. There are further similarities between STaR and expert iteration methods \citep{anthony2017thinking}. For example, 
filtering generated examples based on whether their ultimate answer matches the target can be seen as expert feedback. However, we have a fixed ``expert'' and do not train a separate value function. 

\paragraph{Natural Language Explanations} \looseness=-1
Natural language explanations have also been discussed from the perspective of explainable machine learning, focusing on justification rather than reasoning \citep{camburu2018snli, chen2021generate}. 
The motivation for this line of work is largely grounded in explainable decision making, and similarly to \citet{rajani2019explain}, generally does not find that requiring post-hoc explanations improves model performance. 
\vspace{-10px}
\section{Method}
\vspace{-7px}
\subsection{Rationale Generation Bootstrapping (STaR Without Rationalization)}
\vspace{-3px}

We are given a pretrained LLM $M$ and an initial dataset of problems $x$ with answers $y$: $\mathcal{D} = \{(x_i, y_i)\}_{i = 1}^D$. Our technique starts with a small \emph{prompt} set $\mathcal{P}$ of examples with intermediate \emph{rationales} $r$: $\mathcal{P} = \{(x^p_i, r^p_i, y^p_i)\}_{i = 1}^P$, where $P \ll D$ (e.g.\ $P = 10$). Like standard few-shot prompting, we concatenate this prompt set to each example in $\mathcal{D}$, i.e.\ $x_i = (x^p_1, r^p_1, y^p_1, \dots, x^p_P, r^p_P, y^p_P, x_i)$, which encourages the model to produce a rationale $\hat{r}_i$ for $x_i$ followed by an answer $\hat{y}_i$.
We assume that rationales that lead to correct answers are of better quality than those that lead to incorrect answers. 
Therefore, we filter the generated rationales to include only the ones which result in the correct answer ($\hat{y}_i = y_i$).
We fine-tune the base model $M$ on this filtered dataset, and then restart this process by generating the new rationales with the newly fine-tuned model. We keep repeating this process until the performance plateaus. Note that during this process, once we collect a new dataset, we train from the original pre-trained model $M$ instead of continually training one model to avoid overfitting. We provide an outline of this algorithm in Algorithm~\ref{algostar}.

STaR can be seen as an approximation to an RL-style policy gradient objective. To see this, note that $M$ can be viewed as a discrete latent variable model $p_{M}(y \mid x) = \sum_{r} p(r \mid x) p(y \mid x, r)$; in other words, $M$ first samples a latent rationale $r$ before predicting $y$. Now, given the indicator reward function $\mathbbm{1}(\hat{y} = y)$, the total expected reward across the dataset is
\begin{align}
    J(M, X, Y) &= \sum_i \mathbb{E}_{\hat{r}_i, \hat{y}_i \sim p_{M}(\cdot \mid x_i)} \mathbbm{1}(\hat{y}_i = y_i), \label{eq:rl} \\
    \nabla J(M, X, Y) &= \sum_i \mathbb{E}_{\hat{r}_i, \hat{y}_i \sim p_{M}(\cdot \mid x_i)} \left[ \mathbbm{1}(\hat{y}_i = y_i) \cdot \nabla \log p_M(\hat{y}_i, \hat{r}_i \mid x_i) \right], \label{eq:rlgrad}
\end{align}
where the gradient is obtained via the standard log-derivative trick for policy gradients.
Note that the indicator function discards the gradient for all sampled rationales that do not lead to the correct answer $y_i$: this is the filtering process in STaR (Line 5). Thus, STaR approximates $J$ by (1) greedily decoding samples of $(\hat{r}_i, \hat{y}_i)$ to reduce variance of this estimate (at the cost of potentially biased exploration of rationales), and (2) taking multiple gradient steps on the same batch of data (similar to some policy gradient algorithms \citep{schulman2017proximal}). These approximations make STaR a simple and broadly applicable method that can be implemented with standard LLM training machinery; future work should more closely investigate the link between STaR and the RL objective above.

\subsection{Rationalization}
\begin{wrapfigure}{r}{0.45\textwidth}
\vspace{-30pt}
\lstset{
    escapechar={|}
}
{\small \begin{lstlisting}
Q: Where do you put your grapes just before checking out?
Answer Choices:
(a) mouth
(b) grocery cart |\color{darkspringgreen}(CORRECT)|
(c) super market
(d) fruit basket
(e) fruit market
A: The answer should be the place where grocery items are placed before checking out. Of the above choices, grocery cart makes the most sense for holding grocery items. Therefore, the answer is grocery cart (b).
\end{lstlisting}} \vspace{-10px}
\caption{A few-shot prompt hint we use for rationalization (and not for rationale generation), using the rationale from \citeauthor{wei2022chain}, with its hint included in \textcolor{darkspringgreen}{green}, followed by the rationale and the answer generated by the model. 
}
\vspace{-10px}
\label{fig:rationalization}
\end{wrapfigure}
The rationale generation bootstrapping algorithm 
carries a limitation. 
Since the model is only trained on the examples which it answers correctly, improvement ends when the model fails to solve new problems in the training set. This is fundamentally due to the fact that the algorithm cannot obtain any training signal from failed examples.
Inspired by~\citet{rajani2019explain}, we propose a technique we call ``rationalization''.
Specifically, we provide the answer as a hint to the model and ask it to generate rationales in the same style as in the previous rationale generation step. Given the answer, the model is able to reason backwards, and hence more easily generate a rationale that leads to the correct answer. For example, in Figure~\ref{fig:rationalization}, we provide the hint that ''(b) grocery cart'' is the correct answer in the prompt to generate the rationale. We apply rationalization to the problems which the model failed to solve with rationale generation. When adding a rationalization-generated rationale to our dataset, we do not include the hint in its corresponding prompt, as if the model had come up with the rationale without the hint. 
After filtering, we fine-tune on the previously generated dataset combined with the rationalization-generated dataset. 
\vspace{-5px}
\begin{algorithm}
	\caption{STaR}
	    \hspace*{\algorithmicindent} \textbf{Input} $M$: a pretrained LLM; dataset $\mathcal{D} = \{(x_i, y_i)\}_{i = 1}^D$ (w/ few-shot prompts)
	\begin{algorithmic}[1]
	    \State ${M_0} \leftarrow$ ${M}$ {\color{Gray} \# Copy the original model}
		\For {$n$ \textbf{in} $1...N$} {\color{Gray} \# Outer loop}
		    \State $(\hat{r}_i, \hat{y}_i) \leftarrow M_{n - 1}(x_i)\quad \forall i \in [1, D]$ {\color{Gray} \# Perform rationale generation}
		    {
		    \color{blue}
		    \State $(\hat{r}^\rat_i, \hat{y}^\rat_i) \leftarrow {M_{n - 1}}(\text{add\_hint}(x_i, y_i)) \quad
		    \forall i \in [1, D]$ {\color{Gray} \# Perform rationalization}
		    }
		    \State $\mathcal{D}_n \leftarrow \{ (x_i, \hat{r}_i, y_i) \mid i \in [1, D] \land \hat{y}_i = y_i \}$ {\color{Gray} \# Filter rationales using ground truth answers}
		    {
		    \color{blue}
		    \State $\mathcal{D}^\rat_n \leftarrow \{ (x_i, \hat{r}^\rat_i, y_i) \mid i \in [1, D] \land \hat{y}_i \neq y_i \land \hat{y}^\rat_i = y_i\}$
		    {\color{Gray} \# Filter rationalized rationales}
		    }
		    \State $M_n \leftarrow \text{train}(M, \mathcal{D}_n \, { \color{blue} \cup \, \mathcal{D}^\rat_n } )$ {\color{Gray} \# Finetune the original model on correct solutions - inner loop}
		\EndFor
	\end{algorithmic}
	\label{algostar}
\end{algorithm}
\vspace{-10px}

Algorithm~\ref{algostar} describes the full algorithm, with the parts in {\color{blue} blue} corresponding to rationalization. Without those parts, Algorithm~\ref{algostar} corresponds to STaR without rationalization. Figure~\ref{fig:overview} provides an overview diagram. Fine-tuning on the dataset generated by rationalization has a crucial benefit of exposing the model to difficult problems which otherwise would not have appeared in its finetuning dataset. This can be understood as challenging the model to ``think outside the box'' about problems on which it was unsuccessful. A secondary benefit of rationalization is an increase in dataset size.

\section{Experiments}\label{experimentssection}
\vspace{-7px}
For our experiments, we focus on arithmetic, commonsense reasoning, and grade school math to demonstrate STaR's breadth. In particular, for arithmetic, we follow a setup inspired by \citet{nye2021show}. For commonsense question-answering we follow \citet{xie2021explanation, wei2022chain} and use CommonsenseQA (CQA), a widely used multiple-choice dataset for this domain \citep{talmor2019commonsenseqa}. For grade school math, we use GSM8K from \citet{cobbe2021training}.
\vspace{-4px}
\subsection{Experimental Protocol} \label{expproto}
\vspace{-4px}
We used GPT-J as our base language model, and the fine-tuning script from the GPT-J repository \citep{mesh-transformer-jax}. We chose GPT-J, a 6B-parameter model, because the checkpoint and fine-tuning code are publicly available \citep{mesh-transformer-jax}, and the model is large enough to generate rationales of non-trivial quality to be bootstrapped from. More hyperparameter details about GPT-J and our fine-tuning are included in Appendix~\ref{hyperparameters}. 
Following the default setting of~\citet{mesh-transformer-jax}, we perform a 100-step learning rate warmup, from which point we use a constant learning rate. Unless stated otherwise, we start with $40$ training steps at the first outer loop, and increase the number of fine-tuning training steps by $20\%$ with each outer loop. In general, we found that training more slowly at the beginning ultimately benefits model performance. We expect that further improvement is possible via a thorough hyperparameter search---we leave this to future work due to computational constraints.

For arithmetic problems, we first generate a dataset of 50,000 randomly sampled questions (uniformly over the digit lengths) in the format introduced by~\citet{nye2021show}. For each outer loop iteration on arithmetic, we sample 10,000 problems from the dataset. We use 10 random few-shot rationale examples for each digit for its corresponding few-shot prompt.
For each of the $9,741$ questions in the training set of CommonsenseQA, we add the question to the few-shot rationale prompt, and prompt the model to generate the rationale and answer for that question.
For few shot prompting on CQA, we start with the same 10 questions as used in \citet{wei2022chain}, with the rationales modified slightly to fix an incorrect answer and to more explicitly reference relevant knowledge. We include these modified prompts in Appendix~\ref{modifiedprompts}\footnote{Based on \citet{min2022rethinking}, this is unlikely to meaningfully affect \citeauthor{wei2022chain}'s few-shot performance.}. These prompts serve as our complete set of explanations. We run STaR until we see performance saturate, and we report the best results. 

When performing rationalization, we find that the choice to include or omit few-shot prompts on outer-loop iterations after the first iteration does not have a substantial impact on the method's ultimate performance. However, there are some nuances which we discuss further in Section~\ref{fewshotprompting}, leading us to use few-shot prompts unless stated otherwise. 
\subsection{Datasets}
\paragraph{Arithmetic}

\begin{wrapfigure}{r}{0.3\textwidth} \vspace{-37px}
{
\small
\begin{verbatim}
Input:
6 2 4 + 2 5 9
Target:
<scratch>
6 2 4 + 2 5 9 , C: 0
2 + 5 , 3  C: 1
6 + 2 , 8 3  C: 0
, 8 8 3  C: 0
0 8 8 3 
</scratch>
8 8 3
\end{verbatim}
}\vspace{-5pt}
\caption{A visualization of a 3-digit arithmetic problem with a scratchpad. C corresponds to the carry from the previous digit's summation.}
\vspace{-10pt}
\label{fig:examplescratchpad}
\end{wrapfigure}

The arithmetic task is to calculate the sum of two $n$-digit integers. We generate the dataset based on the descriptions in~\citet{nye2021show} and visualize an example scratchpad in Figure~\ref{fig:examplescratchpad}. Everything up to and including ``\texttt{Target:}'' is given as part of a prompt, and the model is asked to generate the scratchpad (start/end indicated by ``\texttt{<scratch>}'') and the final answer, as in \citet{nye2021show}. Each line of the scratchpad corresponds to the summation of each pair of digits from the final digit to the first digit, the accumulating final digits of the answer, and a carry digit corresponding to whether the previous pair summed to at least 10. We include few-shot prompts for 1 to 5 digits. When performing rationalization, we include the correct answer after ``\texttt{Target}'' and query the model to produce the scratchpad and then reproduce the correct answer following the scratchpad.

\paragraph{CommonsenseQA}
\label{commonsensedataset}
The multiple-choice commonsense reasoning task, CommonsenseQA \citep{talmor2019commonsenseqa} (CQA), is constructed from  ConceptNet, a semantic graph of concepts and their relationships with over a million nodes \citep{speer2016conceptnet}. \citeauthor{talmor2019commonsenseqa} identified a set of ``target'' concepts in ConceptNet for each question, where the target concepts share a semantic relationship to one ``source'' concept. Then each question is crowdsourced to allow a reader to identify one target concept, while mentioning the source concept. In addition, two distractor answers are added. The dataset has 12,247 questions, each with five choices, with 9,741 in the train set, 1,221 in the dev set, and 1,285 in the (withheld) test set. 

Corresponding to the broad variety of ConceptNet, CQA contains a diverse set of questions which require commonsense reasoning ability building off of standard world knowledge, where human performance is 89\% \citep{talmor2019commonsenseqa}. Many have pointed out that CQA contains a number of biases, along several dimensions including gender \citep{rajani2019explain}. We discuss how this may impact our method in Appendix~\ref{biasexplanation}. There are also many typos and questions which are fundamentally ambiguous\footnote{For example, ``Billy bought coffee and waited for his wife to arrive from  France.  Where might he have been?'' includes airport and train station as options. The correct answer, perhaps surprisingly, is train station.}. We use it despite these issues as it is a general question-answering dataset relying on both common world knowledge and simple reasoning, which serves as a good test-bed for our method.

\paragraph{Grade School Math (GSM8K)}
\label{gsmdataset}
We also evaluate on the Grade School Math (GSM8K) dataset, which contains 7,473 train and 1,319 test examples of grade-school-level word problems \citep{cobbe2021training}. These math problems are posed in natural language and require two to eight calculation steps to arrive at a final answer. This dataset combines the skills needed for arithmetic and commonsense reasoning.
\vspace{-5px}

\subsection{Symbolic Reasoning: Results on Arithmetic} \label{arithmeticresults}

\begin{figure}
\centering
\hspace{-40px}
\begin{subfigure}{.46\textwidth}
  \centering
    {
    \small
\hspace{3px}\includegraphics[width=\textwidth]{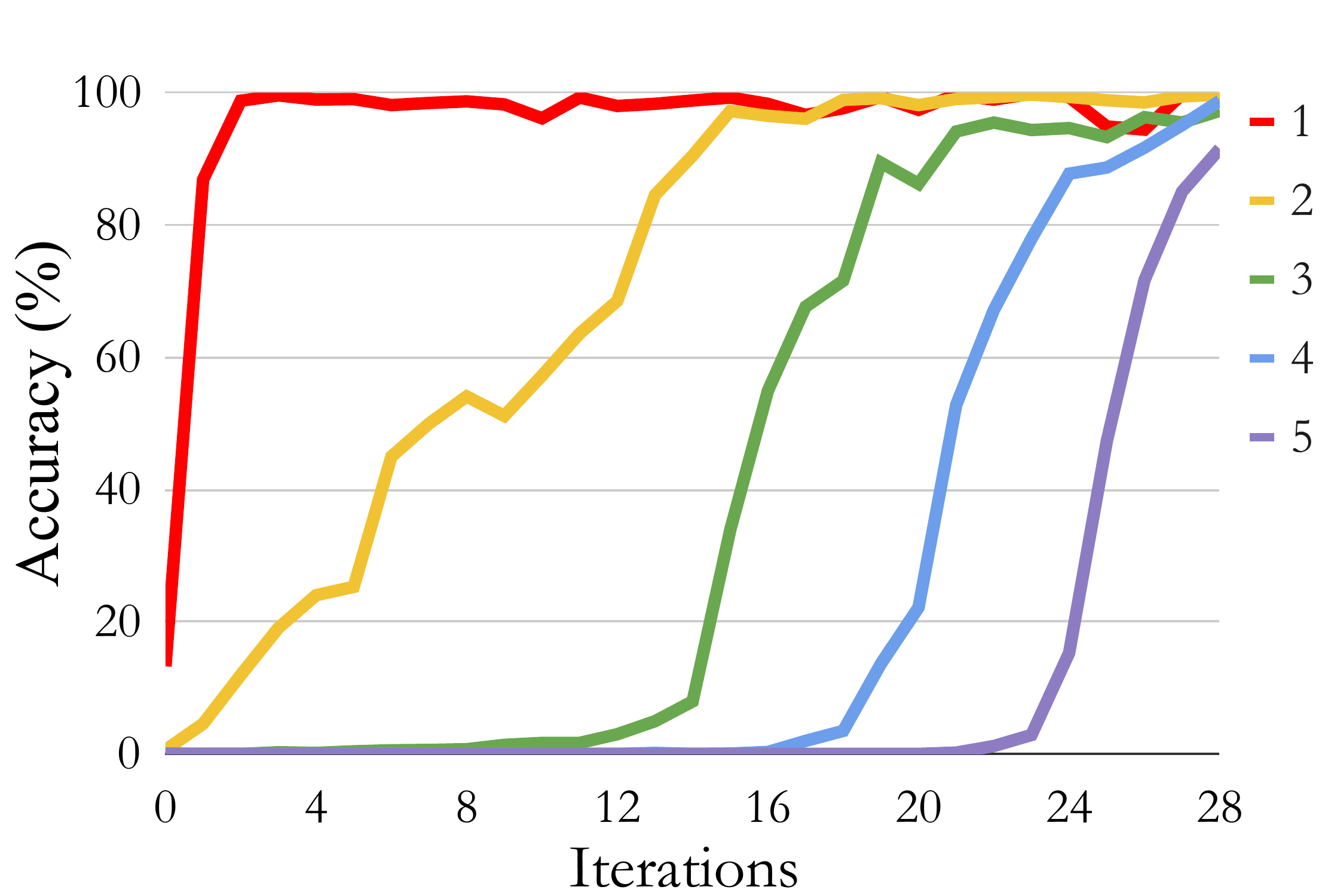}\vspace{-2px} 
\caption{Without rationalization}\vspace{-2px}
}
\end{subfigure}
\begin{subfigure}{.46\textwidth}
  \centering
    {
\hspace{3px}\includegraphics[width=\textwidth]{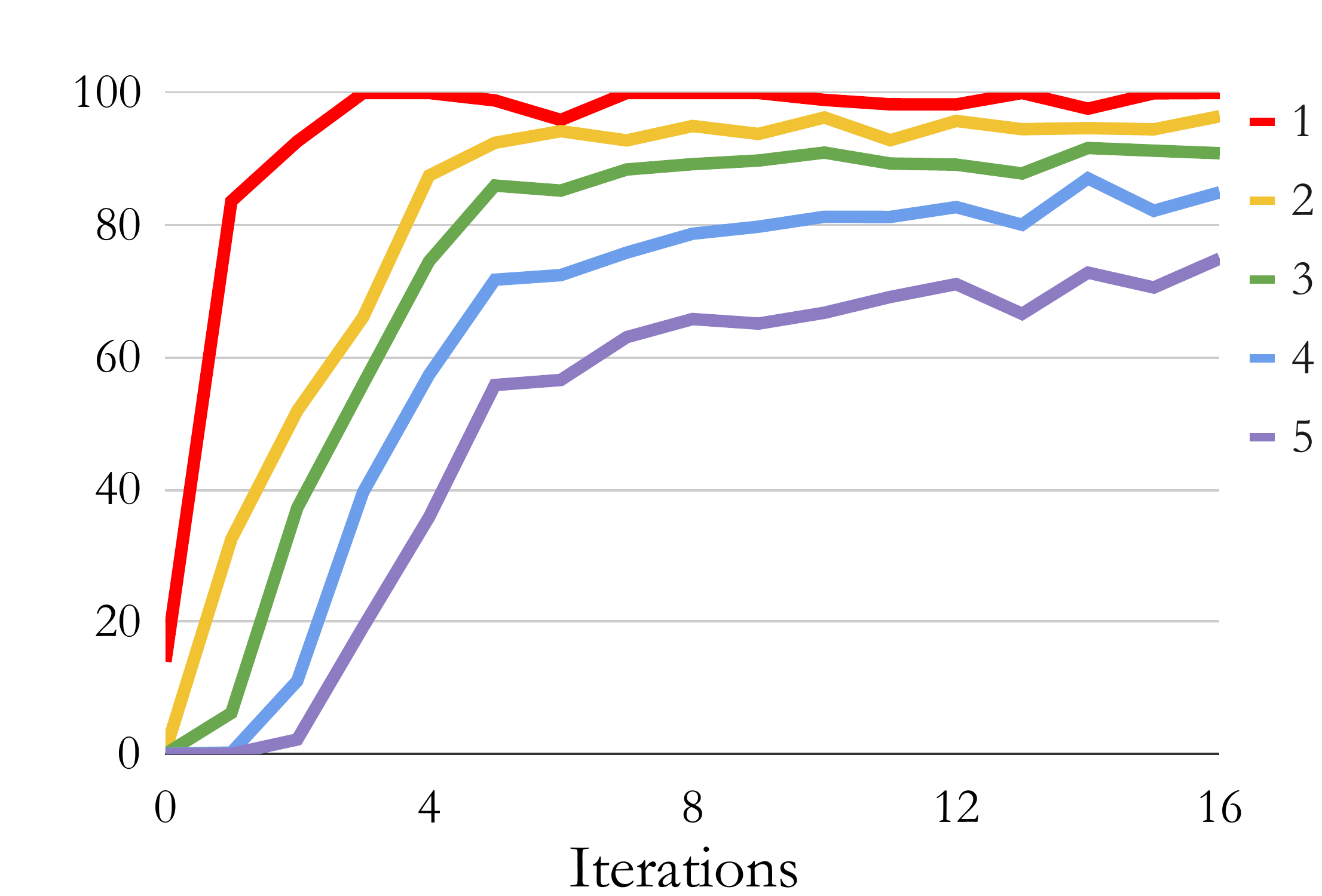}\vspace{-2px}
\caption{With rationalization}\vspace{-2px}
}
\end{subfigure}
\hspace{-40px}
\caption{A visualization of the accuracy of $n$-digit summation with each iteration of STaR with and without rationalization for arithmetic. Each series corresponds to the accuracy of summing two $n$-digit numbers. 
}
\vspace{-15px}
\label{fig:digitschange}
\end{figure}

The accuracies of the model across digits $1$-$5$ over each iteration of the outer loop are plotted in Figure~\ref{fig:digitschange}. After running STaR for 16 iterations, the overall accuracy is $89.5\%$. For reference, a baseline trained on 10,000 examples without rationales for 5,000 steps attains $76.3\%$ accuracy.
Notably, few-shot accuracy on arithmetic problems is very low, even with rationales: accuracy on 2-digit addition is less than $1\%$, and accuracy on more digits close to zero.

\begin{wrapfigure}{r}{0.46\textwidth}
\vspace{-20px}
{
\centering
\hspace{3px}\includegraphics[width=0.46\textwidth]{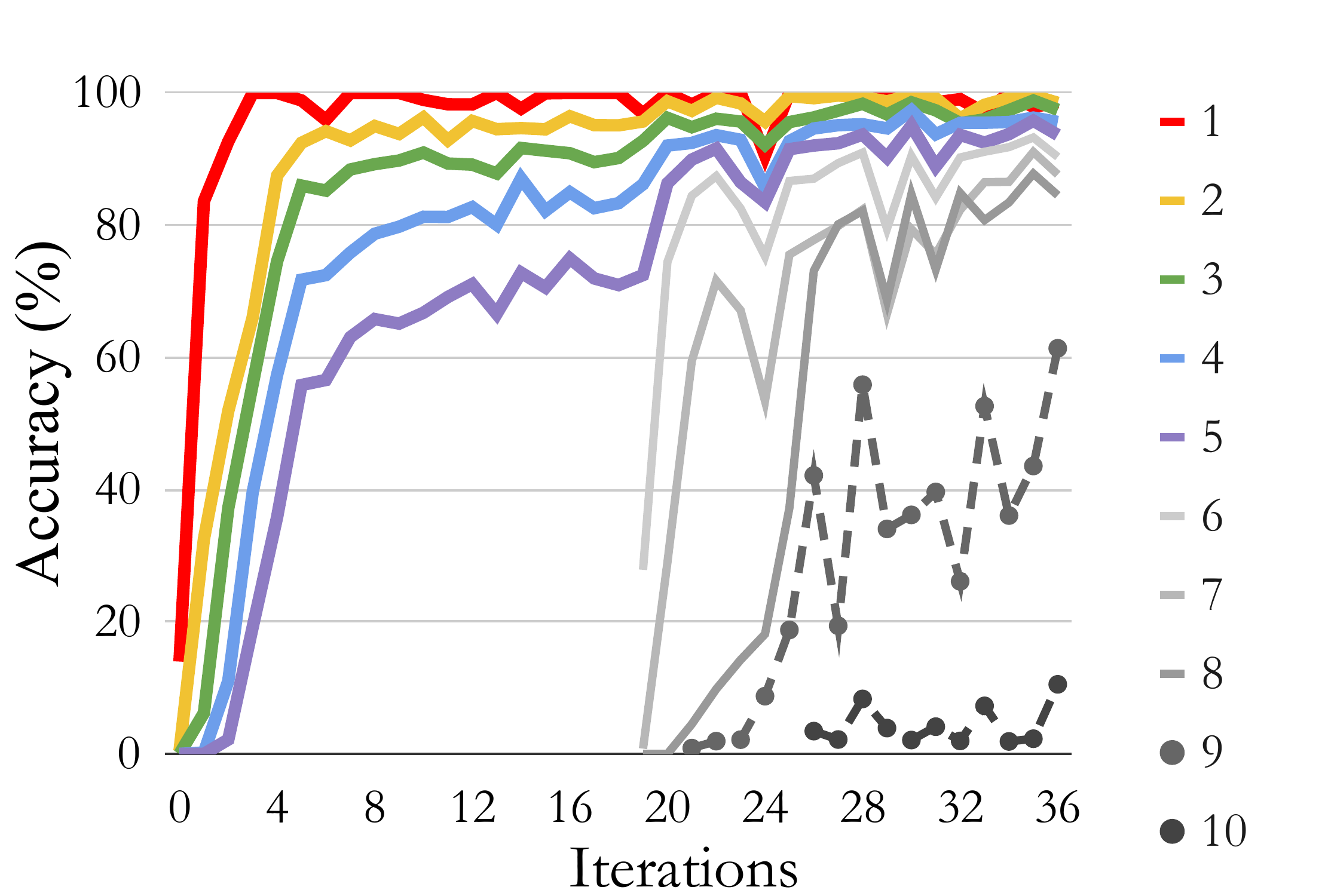}\vspace{-2px} 
}\vspace{-10px}
\caption{We introduce additional digits to STaR with rationalization at the 20$^{th}$ iteration.}
\label{fig:additionaldigits}
\vspace{-10px}
\end{wrapfigure}

With rationalization, the accuracy is able to improve especially quickly.
After one fine-tuning iteration on the model's generated scratchpads, 2-digit addition improves to $32\%$ from less than 1\%.
Without rationalization, the performance improvement is stage-wise: the model generally has poor performance on the $n$-digit sum until it has good performance on the $(n-1)$-digit sum. With rationalization, the model can learn many lengths at once, though not with equal accuracy. 
Rationalization allows many problems to be solved few-shot, so we start STaR training with 300 steps (note, doing so without rationalization causes overfitting on $1$-digit addition), and increase training by 20 steps per iteration. 

We also perform an experiment where we continue pre-training STaR with rationalization with additional digits, starting before the 20th iteration, while keeping the total number of training examples fixed at each iteration. We find that not only does this appear to quickly improve performance on the initial set of digits, but when evaluated on 9 and 10 digit examples, never seen during training, the model successfully solves many of these out-of-distribution problems. As visualized in Figure~\ref{fig:additionaldigits}, the introduction of these digits appears to make the training less stable, but the exact cause is unclear.

\vspace{-5px}\subsection{Natural Language Reasoning: Commonsense Question Answering}\label{commonsenseqaresults}
\vspace{-4px}

The CommonsesenseQA (CQA) setting introduces several new challenges. In the arithmetic task, an incorrect scratchpad in the reasoning step, and to a lesser degree in the rationalization step, was extremely likely to result in an incorrect answer. On the other hand, CQA problems are 5-way multiple choice questions. Thus, one will get the right answer at random approximately 20\% of the time, regardless of the quality of reasoning. Moreover, some simple heuristics (e.g. semantic similarity) can meaningfully improve this to $\approx$30\% without any reasoning, as shown by~\citeauthor{talmor2019commonsenseqa}. 

We evaluate this dataset as described in the experimental protocol and compare to several baselines.
The first baseline is to finetune GPT-J to directly output the final answer, which we call ``GPT-J Finetuned''. We also compare to GPT-3 finetuned to directly predict the final answer from \citet{xu2021human}, and a 137B parameter Lambda model few-shot prompted with chain-of-thought (CoT) rationales from \citet{wei2022chain}.

We found that, as shown in Table~\ref{tab:maintable}, STaR without rationalization outperformed GPT-J fine-tuned directly on the final answer for the entire dataset, despite training on less of the data. The inclusion of rationalization improved this performance to $72.5\%$, far closer to the $73\%$ of the 30$\times$ larger GPT-3. As expected, we also see STaR surpassed the few-shot baselines, including the much-larger 137B LaMDA model \citep{thoppilan2022LaMDA,wei2022chain}. We expect accuracy would be further improved if we applied STaR to a model with higher few-shot performance.
\paragraph{Case Study} Note that it is harder to judge the rationale quality: for arithmetic, one can compare them to the ground truth rationales, but for CQA the evaluation is necessarily qualitative. For this reason, we include a case study in Figure~\ref{fig:exampleresult}. We observe that the rationales provided are generally coherent and of a similar structure to the few-shot rationales. We make the following two observations:\vspace{-5px}
\begin{enumerate}
    \itemsep0em 
    \item After training with STaR, we see the model was able to generate reasonable rationales that solve new problems, which explains part of the observed performance gain.
    \item We also see that there were many instances in which STaR improved the quality of rationales over those generated in a few-shot manner.  
\end{enumerate}\vspace{-5px}
\begin{table}[]
\caption{We evaluate several baselines, including a few-shot GPT-J evaluation both with and without scratchpads, a GPT-J baseline finetuned to directly predict the answer, and STaR with and without rationalization applied to GPT-J. We use CoT to denote non-STaR models outputting rationales, and Direct to indicate those directly predicting the final answer. Note the final STaR model is trained on 78.2\% of the training dataset with rationale generation, and an additional 8.5\% from rationalization.}
\centering
\resizebox{0.8\textwidth}{!}{%
\begin{tabular}{lcc}
                                         & CQA Dev Set Accuracy (\%) & Train Data Used (\%)   \\ \cline{2-3} 
\textit{GPT-3 Direct Finetuned \citep{xu2021human}} & \textit{73.0}      & 100                       \\ \hline
Few-shot Direct GPT-J                           & 20.9               & $\sim$0                   \\
Few-shot CoT GPT-J  \footnotemark               & 36.6               & $\sim$0                   \\
Few-shot CoT LaMDA 137B~\citep{wei2022chain}& 55.6               & $\sim$0                   \\
GPT-J Direct Finetuned                          & 60.0               & 100                       \\
STaR without rationalization             & 68.8               & 69.7                      \\
STaR with rationalization                      & \textbf{72.5}      & 86.7 \\ \hline
\end{tabular}%
}
\vspace{8px}
\label{tab:maintable}
\vspace{-22px}
\end{table}
\footnotetext{We use the same few-shot rationales as described in Section~\ref{expproto} - namely fixing typos and improving clarity.}

\paragraph{Human Evaluation} \label{qualitative}
Based on the observation that STaR may improve reasoning quality for problems even when they were initially answered correctly via few-shot prompting, we performed a preliminary qualitative analysis. We randomly selected 50 rationales generated from few-shot CoT and STaR-generated rationales on questions which they both answered correctly, as well as human-generated rationales for these problems from \citeauthor{rajani2019explain}. We then presented a random subset of 10 questions and rationales to each of 20 crowdworkers on Prolific \citep{palan2018prolific} with the rationales in a randomized order, asking them to rank the rationales based on which they felt best justified the answer. The participants were 30\% more likely to rank the STaR-generated rationales higher than the few-shot rationales ($p=.039$). This indicates that, as mentioned in the case study, STaR can improve the quality of rationale generation.

We also found that the participants were 74\% more likely to prefer the STaR-generated rationales over the human-generated rationales ($p$ < $.001$). To be clear, we do not believe that this indicates human-level rationale-generation performance. Instead, we feel that it speaks to the difficulty of eliciting high-quality rationales. We reproduce the test prompts in Appendix~\ref{humantext} and elaborate on the limitations of the crowdsourced explanations dataset. 

\paragraph{Failure Cases} Finally, we found a variety of interesting failure cases, many of which corresponded to standard logical fallacies. For example, the model often made statements related to the topic of the question but which were not actually arguments for why the answer should be true. Sometimes, the model claimed the question implied the answer as an argument, without explaining why. Other times, especially early in training, the model answered as if it has knowledge about a particular individual, instead of making a general statement - e.g. ``the king's castle is a place where he feels safe'' instead of ``castles are places where kings feel safe.'' We provide examples and analyze errors in Appendix~\ref{errorpatterns}.

\paragraph{Few-shot  Prompt Training}
Including few-shot prompts during fine-tuning~\citep{wei2021finetuned} appears to have a meaningful performance benefit (60.9\% to 68.8\% without rationalization, 69.9\% to 72.5\% with rationalization). Thus, we generally suggest its use for at least some portion of the training, though we discuss some caveats in Section~\ref{fewshotprompting}.

\begin{table}[]
\caption{We find that STaR substantially improves GSM8K performance over the baselines, despite training on only 25.0\% of the data for the model without rationalization, and 28.7\% of the dataset (with 0.5\% from rationalization) for the model with rationalization.}
\centering
\resizebox{0.8 \textwidth}{!}{%
\begin{tabular}{lcc}
                                         & GSM8K Test Accuracy (\%) & Train Data Used (\%)  
\\ \hline
Few-shot Direct GPT-J                           & 3.0               & $\sim$0                   \\
Few-shot CoT GPT-J              & 3.1               & $\sim$0                   \\
GPT-J Direct Finetuned                          & 5.8               & 100                       \\
STaR without rationalization             & 10.1               & 25.0                      \\
STaR with rationalization                                & \textbf{10.7}      & 28.7 \\ \hline
\end{tabular}%
}
\vspace{-10px}
\label{tab:gsmtable}
\end{table}

\begin{wrapfigure}{r}{0.47\textwidth}
\vspace{-20px}
{
\centering
\includegraphics[width=0.47\textwidth]{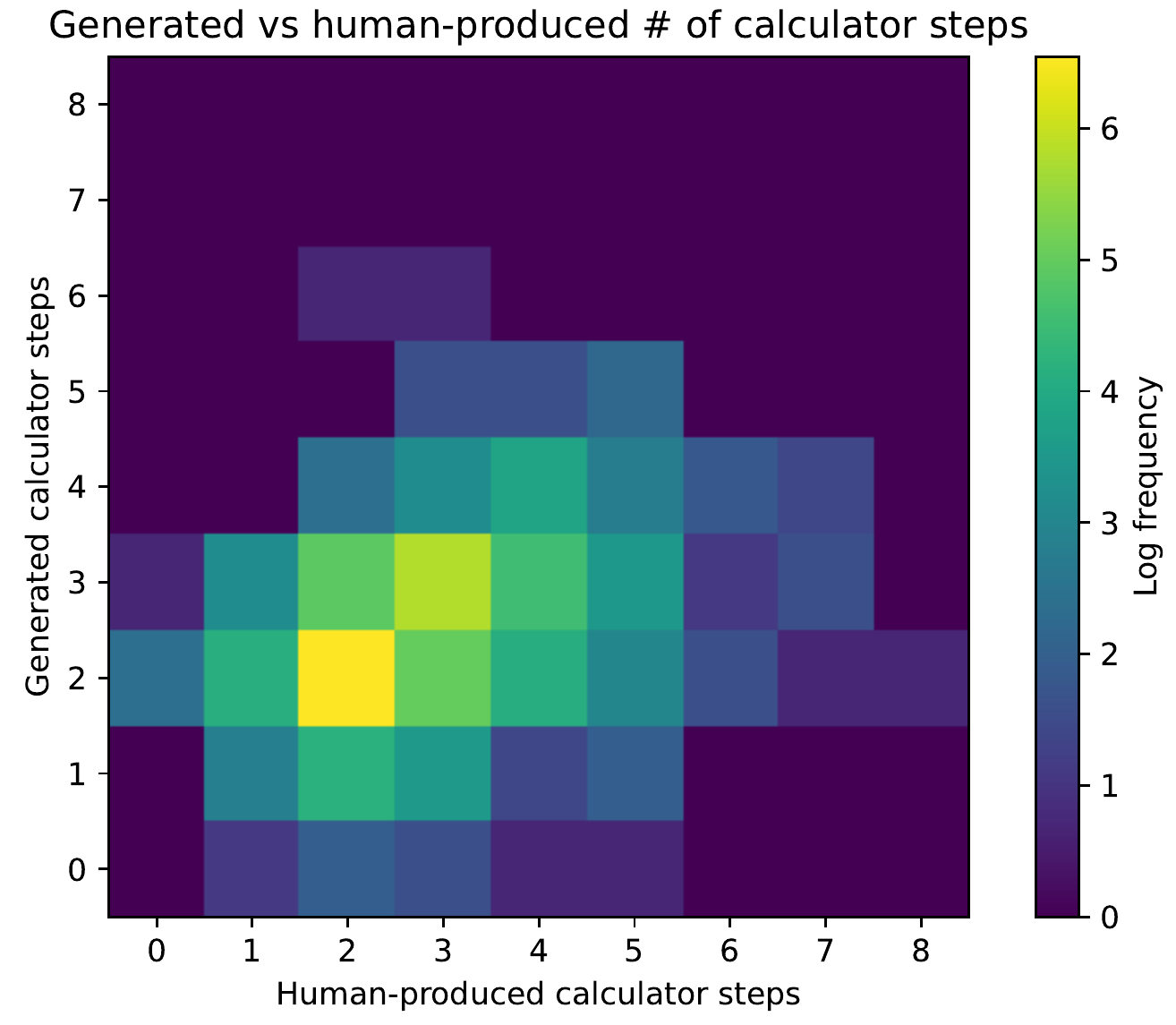}
}\vspace{-10px}
\caption{ A comparison of the number of calculator steps generated by the model in order to solve examples in the training set relative to the number of steps used in the ground truth.}
\label{fig:calculationcorrelation}
\vspace{-20px}
\end{wrapfigure}

\vspace{-5px}
\subsection{Mathematical Reasoning in Language: Grade School Math}\label{gsm8k}

We again find on GSM8K that STaR substantially improves performance beyond few-shot with rationales or training to directly predict the answers (without rationales), shown in Table~\ref{tab:gsmtable} and include the few-shot prompt in Appendix~\ref{gsm8kprompts}. We observe that on this task, the use of rationalization does not substantially improve performance. Note that, in training, it was necessary to cap the number of training steps at the 30th iterations (after 7912 steps), to prevent the training process from becoming prohibitively long. The results were reached after 36 iterations for STaR without rationalization and an additional 10 iterations with rationalization.

Most often, the number of calculation steps generated by the model matches the number of steps taken by humans (generally between 53\% and 57\% agreement across all iterations). We visualize this explicitly in Figure~\ref{fig:calculationcorrelation}. We see that when the ground truth and model disagree on the number of calculation steps, the model typically uses fewer. Sometimes this is because the model skips steps, but occasionally it finds different solutions. We show an example in Appendix~\ref{gsm8ksolutions}, where the model disregards redundant information and solves a 7-step problem in a single step.

\vspace{-8px}
\section{Discussion and Challenges} \label{challenges}
\paragraph{The Impact of Rationalization}
An essential question is exactly what role rationalization plays.
Intuitively, rationalization allows a model to reverse-engineer a solution, or provides a heuristic for identifying whether each step makes the conclusion more likely. This parallels real-world problems where the final result is known, but challenging to derive a good justification.
From a mathematical perspective, while rationale generation samples rationales from the distribution $p(r \mid x)$ provided by our model $M$, rationalization conditions on the answer, letting us access an alternative distribution $p(r \mid x, y)$ which may be a better search space for rationales. Then rationalization could be framed as an off-policy estimate of the objective in Equation~\ref{eq:rl}, sampling from the hint-augmented model as a proposal distribution. Future work should establish more connections between rationalization and these RL objectives, and examine more generally when and why rationalization improves learning. 

In addition, due to the low sampling temperature, the outputs without rationalization correspond to the examples where the model is most confident in its answer. This results in these examples providing a weaker gradient signal than the rationalization examples, at least in the first iteration. Since we retrain from the initial pre-trained model every time we run a fine-tuning iteration, the degree of this effect is also difficult to measure directly. 
Finally, we must point out that the method to add the ``hint'' does not follow immediately from the question and answer and in some contexts providing it may be nontrivial. An exploration of the various impacts of different hinting techniques and their generality is an avenue for future work.

\paragraph{Temperature}
One intuitive alternative to rationalization, if one seeks to expand the training dataset, is more and higher-temperature sampling. However, in practice, we found that this is counterproductive. In general, it substantially increases the likelihood of a correct answer despite incorrect reasoning, and training on bad or irrelevant reasoning prevents generalization. This is particularly clear in more structured tasks, like arithmetic, where the scratchpads that the model learns to produce with a higher-temperature sampling approach diverge into meaninglessness and cause the model to stagnate. Overall, we found that higher temperatures as an alternative to rationalization (e.g. $0.5$ or $0.7$) consistently led to models worse than models with reasoning alone. In addition, as text generation by large language models is sequential (i.e. one cannot produce a token without producing the preceding token), generating text is a bottleneck and this is computationally far less efficient than rationalization. For example, generating 10 sample outputs is approximately 10 times slower than generating one sample output. However, one potentially valuable way to leverage multiple samples would be to use the method proposed in \citet{wang2022selfconsist}, using the majority-vote result of multiple high-temperature scratchpads as a ground truth against which we compare a low-temperature scratchpad. This may allow one to apply STaR to a dataset of only questions, without answers.

\paragraph{Few-shot Prompting}
\label{fewshotprompting}
A noteworthy phenomenon is that the inclusion of few-shot prompting during sampling seems to dramatically reduce ``drift'' where later rationales become increasingly dissimilar from the initial few-shot set of rationales. One benefit of this is that the model may be less constrained by the quality and difficulty of the initial rationales, theoretically allowing it to generalize more. One potentially negative consequence is that the style of the rationales may less-closely match the original prompting style. Another benefit is in terms of computational resources - a shorter prompt length allows for a shorter sequence length when sampling. Technically, the point in training at which we ``disable'' few-shot prompts is another hyperparameter which we could tune, but we leave this to future work. In addition, by leaving prompts out after the initial outer-loop iteration, the model tends to perform gradually worse at rationalization as it trains for longer periods of time. As a result, it may be necessary to include some hints during training for long periods of time with this approach. 

Ultimately, the choice to include few-shot prompts in later iterations of training appears to depend on the use-case: when the goal is consistent adherence to a particular prompt style, which may benefit explainability, include few-shot prompts in sampling; when the goal is a faster training loop, one may remove them. Moreover, it is possible that with other datasets or larger models there is an impact on performance, so we encourage this to be generally treated as a hyperparameter. 
\vspace{-5px}
\section{Conclusion} \label{conclusion}\vspace{-2px}
We present the Self-Taught Reasoner (STaR), which iteratively improves a model's ability to generate rationales to solve problems. 
We few-shot prompt a model to solve many problems in a step-by-step manner by generating rationales, and then prompt it to rationalize the correct answer for problems it gets wrong. We finetune on both the initially correct solutions and rationalized correct solutions, and repeat the process.
We find that this technique significantly improves the model's generalization performance on both symbolic reasoning and natural language reasoning.

There are several important limitations on STaR as presented. In order for the first iteration of STaR to succeed, few-shot performance must be above chance, implying that the initial model must be big enough to have some reasoning capabilities. For instance we found that GPT-2 was not able to bootstrap from few-shot reasoning in even the arithmetic domain. A further limitation is that settings with a high level of chance performance (e.g.~binary decisions) yield many poor rationales, confounding the STaR approach. An open problem is how to filter bad reasoning in these settings.

\looseness=-1
Nonetheless, we believe using examples without reasoning to bootstrap reasoning is a very general approach, and that STaR can serve as the basis of more sophisticated techniques across many domains.

\section*{Acknowledgements}
We thank Imanol Schlag for his detailed feedback about this work, as well as Rose E Wang, Markus Rabe, Aitor Lewkowycz, Rishi Bommasani, Allen Nie, Alex Tamkin, and Qian Huang. We thank Cem Anil for his very helpful insight that rationale finetuning performance can be improved if the training includes the few-shot rationales. We also thank Ben Prystawski for his suggestions on survey creation. We thank Google TPU Research Cloud for TPU access.

\bibliographystyle{unsrt}
\bibliography{refs}

\vfill
\pagebreak


\appendix

\newpage
\section*{Appendix}
\section{CommonsenseQA Error Patterns}
\label{errorpatterns}
Throughout our experiments, we came across a variety of interesting failure cases for commonsense reasoning. Note that all the final answers are correct -- however, we take issue with the reasoning used in order to arrive at those answers.

\subsection{Question Implies Answer}
One key failure case was answers in the form of ``the answer must be something that is <question property>. <answer> is <question property>. Therefore, the correct answer is <answer>.'' In these cases, the model fails to explain \textit{why} the answer that it has chosen satisfies the question property. 

\begin{verbatim}
Q: What might someone get from learning about science?
Answer Choices:
(a) headache
(b) see things differently
(c) increased knowledge
(d) accidents
(e) appreciation of nature
A: The answer must be something that someone would get from learning about 
science. Learning about science would increase knowledge. Therefore, the 
answer is increased knowledge (c).
\end{verbatim}

These rationales, while perhaps useful to the model, read to us as opaque and unexplanatory. 

\subsection{Begging the Question}
A related but stronger version of the previous failure case, while less common, is particularly uninsightful. Sometimes the model will imply the answer that it has chosen in its question.

\begin{verbatim}
Q: The christian rock band was on a large stage, the singer's foot got caught
on a wire and he had a small what?
Answer Choices:
(a) commit sin
(b) fail
(c) disobey
(d) stumble
(e) boo boo
A: The answer must be something that would cause the singer to stumble. The 
singer's foot got caught on a wire, so he would stumble. Therefore, the 
answer is stumble (d).
\end{verbatim}

\subsection{Exercise to the Reader}
A rare failure case is when the model finds it unnecessary to justify its answer. For example:

\begin{verbatim}
Q: Other than school or from text where can one gain knowledge?
Answer Choices:
(a) book
(b) class
(c) thinking
(d) head
(e) meeting
A: The answer must be a place where someone can gain knowledge other than 
school or from text. The answer is meeting (e).
\end{verbatim}

\subsection{World State Assertions}
Sometimes, the model will assume that it knows something about a subject or a person whose name was used as a variable. This leads to somewhat comical examples of reasoning. Part of the reason for this is that generally, there is an expectation that good rationales will leverage understanding of more general classes of objects and appeal to the relationship between those general classes and the particular instance. For example, the argument that ``a person would typically feel exhilaration from heights'' is generally more compelling than the argument that ``James would feel exhilaration from heights.''

\begin{verbatim}
Q: A person's hand hurts after hitting the bottle. What might happen to 
the bottle?
Answer Choices:
(a) alcoholism
(b) breaking
(c) fly
(d) nausea
(e) ticket
A: The answer must be something that happens to the bottle. The bottle is
broken. Therefore, the answer is breaking (b).
\end{verbatim}

\begin{verbatim}
Q: James loved climbing.  What emotion might he feel from heights?
Answer Choices:
(a) exhaustion
(b) exhilaration
(c) falling down
(d) nervousness
(e) vertigo
A: The answer must be a feeling that would cause someone to climb a mountain.
James would feel exhilaration from heights. Therefore, the answer is 
exhilaration (b).
\end{verbatim}

\subsection{Red Herrings}
Some errors in reasoning corresponded to the model making a statement which, while technically true, is not useful in demonstrating the claim. 

\begin{verbatim}
Q: A squirrel ran up to it's home, where is the small dog likely to be
barking up?
Answer Choices:
(a) manhattan apartment
(b) wood
(c) neighbor's yard
(d) tree
(e) basket
A: The answer must be a place where a small dog would be likely to be
barking. Squirrels are not small dogs. Therefore, the answer is tree (d).
\end{verbatim}

\begin{verbatim}
Q: John ran very fast, his feet touched the ground in what way?
Answer Choices:
(a) behind
(b) loose
(c) feverishly
(d) slowly
(e) lightly
A: The answer must be a way of running that is different from the way John 
ran. Running is a fast, light-footed activity. Therefore, the answer is 
lightly (e).
\end{verbatim}

\subsection{Hint Short-cutting}
In the experiments where the model was shown some examples of ``hints'' during training, in order to prevent it from losing the ability to perform rationalization over time, the model appeared to pick up on the fact that the final answer would always correspond to the hinted answer. This led to answers such as 

\begin{verbatim}
Q: Billy bought coffee and waited for his wife to arrive from  France. 
Where might he have been?
Answer Choices:
(a) airport
(b) grocery store
(c) internet cafe
(d) supermarket
(e) train station
A: The answer must be a place where Billy could have been waiting for his
wife to arrive from France. The airport is a place where people can wait 
for flights. Therefore, the answer is train station (e).
\end{verbatim}

\newpage
\section{Modified CQA Prompts}
\label{modifiedprompts}
For reference, we include our modified prompts based closely on those in \citet{wei2022chain}.

\begin{lstlisting}
Q: What do people use to absorb extra ink from a fountain pen?
Answer Choices:
(a) shirt pocket
(b) calligrapher's hand
(c) inkwell
(d) desk drawer
(e) blotter
A: The answer must be used to absorb extra ink. Blotters are designed to absorb liquids. Therefore, the answer is blotter (e).

Q: What home entertainment equipment requires cable?
Answer Choices:
(a) radio shack
(b) substation
(c) television
(d) cabinet
(e) desk
A: The answer must require cable. Cable is used to provide satellite channels to televisions. Therefore, the answer is television (c).

Q: The fox walked from the city into the forest, what was it looking for?
Answer Choices:
(a) pretty flowers
(b) hen house
(c) natural habitat
(d) storybook
(e) dense forest
A: The answer must be a reason for a fox to go into the forest. The forest is a fox's natural habitat. Therefore, the answer is natural habitat (c).

Q: Sammy wanted to go to where the people were. Where might he go?
Answer Choices:
(a) populated areas
(b) race track
(c) desert
(d) apartment
(e) roadblock
A: The answer must be a place with many people. Populated areas, by definition, have a lot of people. Therefore, the answer is populated areas (a).

Q: Where do you put your grapes just before checking out?
Answer Choices:
(a) mouth
(b) grocery cart
(c) super market
(d) fruit basket
(e) fruit market
A: The answer should be the place where grocery items are placed before checking out. Of the above choices, grocery cart makes the most sense for holding grocery items. Therefore, the answer is grocery cart (b).

Q: Google Maps and other highway and street GPS services have replaced what?
Answer Choices:
(a) united states
(b) mexico
(c) countryside
(d) atlas
(e) oceans
A: The answer must be something that used to do what Google Maps and GPS services do, which is give directions. Atlases were also used to give directions. Therefore, the answer is atlas (d).

Q: Before getting a divorce, what did the wife feel who was doing all the work?
Answer Choices:
(a) harder
(b) anguish
(c) bitterness
(d) tears
(e) sadness
A: The answer should be a feeling which would cause someone who was doing all the work to get divorced. If someone feels bitter towards their spouse, they are likely to want a divorce. Therefore, the answer is bitterness (c).
\end{lstlisting}

\newpage
\section{Human-evaluated Test Prompts}
\label{humantext}
We also selected a random sampling of 50 questions which were correctly answered both few-shot and by a STaR-trained model (without rationalization), as discussed in \ref{qualitative}. Presented in a random order, twenty crowdworkers preferred the STaR-generated answers. We reproduce the examples here with the few-shot rationale first, the STaR-trained rationale second, and the human rationale third, though these were shuffled when presented to participants. We selected human answers from \citeauthor{rajani2019explain}'s original split rationales where possible, finding that duplicate rationales were much more common in the new split rationales. For example, the explanation ``Rivers flow trough valleys,'' appeared over 400 times verbatim in the new split dataset, and ``This word was most relevant'' appeared over 150 times. \citeauthor{rajani2019explain}'s dataset also includes explanations like ``The only answer that makes sense'' or ``BOB WILL NOT POKEMON CARDS WERE COMMON AND WORTHLESS BUT WRONG ABOUT THEM SO FEEL REALLY RARE TO DELAY'' or restatements of the answer. We append the phrase ``Therefore, the answer is ANSWERTEXT (ANSWERLETTER)'' with ANSWERTEXT replaced by the correct answer's text and ANSWERLETTER replaced by the correct answer letter. This is done 1) to make it less obvious that one of the answers is generated by a different source and 2) to prioritize differences in rationales, not the answer format. 

Before the questions and after the consent form, we presented the following directions:

\begin{lstlisting}[basicstyle=\tiny\ttfamily,
columns=flexible,
breaklines=true,
breakindent=0pt]On the next (and only) survey page, you will be presented with 10 questions.

Each question will have three explanations for one answer, which is a pre-determined correct ("best") answer.

Please rank the explanations for each question from best to worst (with 1 as the best and 3 as the worst) based on how well they justify the correct answer.

Note there are some typos and grammatical errors in the examples, in the questions, answers, and the explanation. Please treat the answer letter (a / b / c / d / e) at the end of the explanations as the correct answer letter, even if another answer is valid (and perhaps more reasonable), and even if it does not match with the answer text preceding it.
\end{lstlisting}
The examples were subsampled and presented to the crowdworkers:
{
\begin{multicols}{2}
\begin{lstlisting}[basicstyle=\tiny\ttfamily,
columns=flexible,
breaklines=true,
breakindent=0pt]
Q: Who is likely yo have a caring heart?
Answer Choices:
(a) person
(b) species
(c) artichoke
(d) all mammals
(e) live animal
 A: The answer must be a person. All mammals have a caring heart. Therefore, the answer is person (a).
 A: The answer must be someone who is likely to have a caring heart. Humans are the most likely to have a caring heart. Therefore, the answer is person (a).
 A: A person is the only human being capable of having a caring heart. Therefore, the answer is person (a).

Q: If a person wanted to become a good writer what should they do in college?
Answer Choices:
(a) word sentence
(b) own animal
(c) read the newspaper
(d) catch cold
(e) study literature
 A: The answer must be something that would help a person become a good writer. The answer is study literature (e).
 A: The answer must be something that would help a person become a good writer. Literature is a form of study. Therefore, the answer is study literature (e).
 A: studying literature will give one skills of writing. Therefore, the answer is study literature (e).

Q: The human always would have fun making up questions for the A.I. overlords, he found the task quite what?
Answer Choices:
(a) do enjoy
(b) eat cake
(c) enjoy living
(d) get laid
(e) enjoyable

 A: The answer must be something that the human would enjoy doing. The human enjoys making up questions for the A.I. overlords. Therefore, the answer is enjoyable (e).
 A: The answer must be something that the human found enjoyable. The human found the task of making up questions for the A.I. overlords enjoyable. Therefore, the answer is enjoyable (e).
 A: Having fun is enjoyable. Therefore, the answer is enjoyable (e).



Q: What happens when I make a pig of myself eating breakfast?
Answer Choices:
(a) full stomach
(b) gain energy
(c) dieting
(d) have energy
(e) feeling satisfied

 A: The answer must be a feeling which would cause someone to feel full after eating breakfast. The answer is full stomach (a).
 A: The answer must be a feeling which would cause someone to make a pig of themselves eating breakfast. If someone feels full, they are likely to stop eating. Therefore, the answer is full stomach (a).
 A: A colloquial phrase for over indulging in food. Therefore, the answer is full stomach (a).

Q: In order to see a story on the big screen what must you do?
Answer Choices:
(a) go to movies
(b) visualize
(c) reading
(d) open book
(e) sketching a picture

 A: The answer must be something that you do to see a story on the big screen. Movies are a way to see a story on the big screen. Therefore, the answer is movies (a).
 A: The answer must be something that is required to see a story on the big screen. Movies are a form of storytelling. Therefore, the answer is go to movies (a).
 A: The big screen is a colloquial term meaning a movie theater. Therefore, the answer is go to movies (a).

Q: What is the opposite of remembering?
Answer Choices:
(a) answering question
(b) stopping to think
(c) knowing
(d) closing eyes
(e) forgetting
 A: The answer must be something that is the opposite of remembering. The answer is forgetting (e).
 A: The answer must be a feeling which would cause someone to not remember something. If someone forgets something, they are likely to not remember it. Therefore, the answer is forgetting (e).
 A: Health complications. Therefore, the answer is forgetting (e).

Q: If you aren't experiencing curiosity and want to ensure viewing enjoyment, what may you watch?
Answer Choices:
(a) examine thing
(b) stare into a mirror
(c) go somewhere
(d) hear news
(e) see favorite show
 A: The answer must be something that will make you want to watch. If you are not experiencing curiosity, you may want to watch something that will make you want to watch. Therefore, the answer is watch (e).
 A: The answer must be something that will not cause you to want to watch. If you are not experiencing curiosity, you are likely to want to watch something that is entertaining. Therefore, the answer is favorite show (e).
 A: see favorite show can you do in front of a tv if you have curiosity about what is being displayed on the tv. Therefore, the answer is favorite show (e).

Q: The carotid is a main artery susceptible to damage where?
Answer Choices:
(a) neck
(b) face
(c) busy city
(d) torso
(e) human body
 A: The answer must be a place where the carotid artery is susceptible to damage. The carotid artery is located in the neck. Therefore, the answer is neck (a).
 A: The answer must be a place where the carotid is susceptible to damage. The carotid is susceptible to damage in the neck. Therefore, the answer is neck (a).
 A: The carotid artery is located in the neck. Therefore, the answer is neck (a).

Q: The king needed to feel safe, where did he go?
Answer Choices:
(a) castle
(b) throne room
(c) deck of cards
(d) fort
(e) court
 A: The answer must be a place where the king feels safe. The king's castle is a place where he feels safe. Therefore, the answer is castle (a).
 A: The answer must be a place where the king can feel safe. Castles are places where kings feel safe. Therefore, the answer is castle (a).
 A: In general, King live in castle. Therefore, the answer is castle (a).

Q: What is the result of instituting civil action?
Answer Choices:
(a) bad feelings
(b) confusion
(c) going to court
(d) hurt feelings
(e) sandwiches
 A: The answer must be a result of instituting civil action. The result of instituting civil action is a lawsuit. Therefore, the answer is going to court (c).
 A: The answer must be a result of instituting civil action. If someone goes to court, they are likely to feel hurt feelings. Therefore, the answer is going to court (c).
 A: Anger and Trial will not institute a court case. Therefore, the answer is going to court (c).Q: The person wanted to fix his car, what did he order?
Answer Choices:
(a) selling things
(b) gas
(c) feel resentful
(d) machine part
(e) cross street
 A: The answer must be something that would help the person fix his car. The person would not want to sell things, because he would not want to give up his car. Therefore, the answer is machine part (d).
 A: The answer must be something that would be used to fix a car. Machine parts are used to fix cars. Therefore, the answer is machine part (d).
 A: A machine part is the only thing listed that would fix the car. Therefore, the answer is machine part (d).

Q: He always traveled up the mountain when winter arrived, he had a passion and desire to do what?
Answer Choices:
(a) play volleyball
(b) play hockey
(c) ski
(d) knit
(e) skate
 A: The answer must be something that the person does when winter arrives. The answer is skiing (c).
 A: The answer must be something that the person did when winter arrived. The answer is ski (c).
 A: Skiing is the only option that is done on a mountain. Therefore, the answer is ski (c).

Q: Bob thought that his Pokemon cards were common and worthless, but he was wrong about them. They were really what?
Answer Choices:
(a) personal
(b) special
(c) shiny
(d) rare
(e) irregular
 A: The answer must be something that is not common and worthless. Pokemon cards are not common and worthless. Therefore, the answer is rare (d).
 A: The answer must be something that are rare. Pokemon cards are rare. Therefore, the answer is rare (d).
 A: BOB WILL NOT POKEMON CARDS WERE COMMON AND WORTHLESS BUT WRONG ABOUT THEM SO FEEL REALLY RARE TO DELAY. Therefore, the answer is rare (d).

Q: If you're lighting a candle to see what are you in?
Answer Choices:
(a) dimly lit room
(b) synagogue
(c) sunlight
(d) birthday cake
(e) bedroom
 A: The answer must be a place where you can see what you are in. The answer is a room with dim lighting. Therefore, the answer is dimly lit room (a).
 A: The answer must be a place where you can see a candle. Candles are used to see things in the dark. Therefore, the answer is dimly lit room (a).
 A: Lighting or illumination is the deliberate use of light to achieve a practical or aesthetic effect. Lighting includes the use of both artificial light sources like lamps ... Therefore, the answer is dimly lit room (a).

Q: The person was thirsty, what should he have done?
Answer Choices:
(a) eat
(b) sunny days
(c) compliments
(d) drink water
(e) all kinds of things
 A: The answer must be something that would make the person feel better. If the person is thirsty, he should drink water. Therefore, the answer is drink water (d).
 A: The answer must be something that would make someone feel better if they were thirsty. If someone is thirsty, they are likely to want to drink water. Therefore, the answer is drink water (d).
 A: The person was thirsty, so drink water. Therefore, the answer is drink water (d).

Q: Where would you find many products for sale?
Answer Choices:
(a) market
(b) cost one dollar
(c) shelf
(d) salon
(e) demonstration
 A: The answer must be a place where many products are sold. A market is a place where many products are sold. Therefore, the answer is market (a).
 A: The answer must be a place with a lot of products for sale. Markets are the most common place to buy products. Therefore, the answer is market (a).
 A: The only place products are for sale are in markets and stores. Therefore, the answer is market (a).

Q: If man and mammoth wanted to meet, they would probably have to do it during when?
Answer Choices:
(a) a baby shower
(b) ancient history
(c) storybook
(d) stone age
(e) forest
 A: The answer must be a time when man and mammoth would meet. Stone age is the time when man and mammoth lived. Therefore, the answer is stone age (d).
 A: The answer must be a time when man and mammoth would be able to meet. The stone age was a time when man and mammoth would be able to meet. Therefore, the answer is stone age (d).
 A: During the stone age is the only time mammoths lived. Therefore, the answer is stone age (d).

Q: They prided themselves on being a wealth of knowledge, and that's why many chose to attend their what?
Answer Choices:
(a) book store
(b) university
(c) meeting
(d) class
(e) encyclopedia
 A: The answer must be a place where people go to learn about things. The answer is university (b).
 A: The answer must be a place where people go to learn about things. Universities are places where people go to learn about things. Therefore, the answer is university (b).
 A: because in the university you get more knowledge. Therefore, the answer is university (b).

Q: what happens when visiting museum that have narrow halls?
Answer Choices:
(a) sore feet
(b) amazement
(c) retrospection
(d) boredom
(e) crowding
 A: The answer must be something that happens when visiting a museum that has narrow halls. Crowding is a common problem in museums. Therefore, the answer is crowding (e).
 A: The answer must be something that would cause someone to have sore feet if they are walking in narrow halls. If someone feels cramped, they are likely to want to go somewhere else. Therefore, the answer is crowding (e).
 A: museum is one of the crowdest place. Therefore, the answer is crowding (e).

Q: On their hike they brought a filtering straw, they were worried about germs in the what?
Answer Choices:
(a) make sick
(b) doctor
(c) water
(d) stream
(e) mouth
 A: The answer must be something that can filter out germs. Filtering straws are used to filter out germs. Therefore, the answer is filtering straw (c).
 A: The answer must be something that would cause someone to bring a filtering straw on a hike. Filtering straws are used to filter water. Therefore, the answer is water (c).
 A: straw it something used to drink water. Therefore, the answer is water (c).

Q: If I am not good at playing poker, I will begin what if I play?
Answer Choices:
(a) discord
(b) social event
(c) losing money
(d) competition
(e) heightened awareness
 A: The answer must be something that will happen if you are not good at playing poker. If you are not good at playing poker, you will begin to lose money. Therefore, the answer is losing money (c).
 A: The answer must be something that will help you if you are not good at playing poker. If you are losing money, you are likely to want to stop playing. Therefore, the answer is losing money (c).
 A: Poker is a game in which people lose their money. Therefore, the answer is losing money (c).

Q: People are meeting people, what is the likely physical gesture?
Answer Choices:
(a) introducing yourself
(b) smiling
(c) shake hands
(d) share a hug
(e) talking
 A: The answer must be a physical gesture that is used to greet someone. The answer is handshake (c).
 A: The answer must be a gesture used to meet people. If someone wants to meet people, they are likely to want to shake hands. Therefore, the answer is shaking hands (c).
 A: A handshake is a globally widespread. Therefore, the answer is shaking hands (c).

Q: What do dogs get that make them itch?
Answer Choices:
(a) bones
(b) fleas
(c) bad breath
(d) wagging tails
(e) four legs
 A: The answer must be something that makes dogs itch. Dogs get fleas, which make them itch. Therefore, the answer is fleas (b).
 A: The answer must be something that dogs get that make them itch. Fleas are a common cause of dog itch. Therefore, the answer is fleas (b).
 A: fleas makes you itch. Therefore, the answer is fleas (b).

Q: What is ignorance often equated with?
Answer Choices:
(a) bliss
(b) hostility
(c) peace
(d) accidents
(e) find truth
 A: The answer must be something that is often equated with ignorance. Bliss is often equated with ignorance. Therefore, the answer is bliss (a).
 A: The answer must be a feeling which would cause someone to be ignorant. If someone feels bliss, they are likely to be ignorant. Therefore, the answer is bliss (a).
 A: bliss comes from ignorance. Therefore, the answer is bliss (a).

Q: The two teams were competing all the time, what likely developed between them?
Answer Choices:
(a) accomplishment
(b) enemies
(c) rivalry
(d) friendship
(e) bad blood
 A: The answer must be something that developed between the two teams. The teams were competing all the time. Therefore, the answer is rivalry (c).
 A: The answer must be a feeling which would cause two teams to compete all the time. Rivalry is a common feeling which would cause two teams to compete all the time. Therefore, the answer is rivalry (c).
 A: likely developed between them of rivalry. Therefore, the answer is rivalry (c).

Q: James doesn't want a relationship. He thinks that loving another will only create what?
Answer Choices:
(a) satisfaction
(b) jealousy
(c) happiness
(d) wonderful
(e) quiet
 A: The answer must be something that would make James feel bad about not having a relationship. If someone feels jealous, they are likely to want a relationship. Therefore, the answer is jealousy (b).
 A: The answer must be a feeling which would cause someone who doesn't want a relationship to not want one. If someone feels jealous of another, they are likely to not want a relationship. Therefore, the answer is jealousy (b).
 A: Jealousy is a negative emotion. Therefore, the answer is jealousy (b).

Q: The had a reputation for celebrating in a  debaucherous manner, what often happened as a result?
Answer Choices:
(a) drunkenness
(b) enjoyment
(c) have fun
(d) hang over
(e) fighting
 A: The answer must be a result of a celebration. The answer is drunkenness (a).
 A: The answer must be something that would cause someone to have a reputation for celebrating in a debaucherous manner. Alcohol is a depressant. Therefore, the answer is drunkenness (a).
 A: it is normal not a good idea to get drunk. Therefore, the answer is drunkenness (a).

Q: The paper towels are usually in plain sight in the kitchen, where did the dad go next to find them?
Answer Choices:
(a) pantry
(b) cabinet
(c) grocery store
(d) waste bin
(e) market
 A: The answer must be a place where the dad would go to find the paper towels. The pantry is a place where food is stored. Therefore, the answer is pantry (a).
 A: The answer must be a place where paper towels are stored. Pantries are the most likely place to store paper towels. Therefore, the answer is pantry (a).
 A: The pantry is in the kitchen and stores things. Therefore, the answer is pantry (a).

Q: My house is very dry, with little water in the air. What tool should I use to fix this?
Answer Choices:
(a) humidifier
(b) surface of earth
(c) skating rink
(d) ice cube
(e) burn it to the ground
 A: The answer must be a tool that will make the air in your house moist. A humidifier is a tool that will make the air in your house moist. Therefore, the answer is humidifier (a).
 A: The answer must be something that would make the air in my house more humid. A humidifier is used to make the air in a room more humid. Therefore, the answer is humidifier (a).
 A: This word was most relevant. Therefore, the answer is humidifier (a).

Q: A motorcycle is quicker because it isn't as what as a car?
Answer Choices:
(a) heavy
(b) shiny
(c) red
(d) obesity
(e) expensive
 A: The answer must be something that is not as heavy as a car. A motorcycle is lighter than a car. Therefore, the answer is not heavy (a).
 A: The answer must be something that is not as quick as a motorcycle. Motorcycles are not as heavy as cars. Therefore, the answer is heavy (a).
 A: Because it is much smaller. Therefore, the answer is heavy (a).

Q: The baseball team was playing ball. It was time for the pitcher to act. How did the pitcher get the ball to the batter?
Answer Choices:
(a) lose
(b) injury
(c) competition
(d) throwing
(e) rolled
 A: The answer must be a way to get the ball to the batter. The pitcher is the one who throws the ball. Therefore, the answer is throwing (d).
 A: The answer must be a way for the pitcher to get the ball to the batter. The pitcher is the person who is responsible for getting the ball to the batter. Therefore, the answer is throwing (d).
 A: The pitcher threw the ball to the batter so that he could hit it. Therefore, the answer is throwing (d).

Q: In what sort of container should you keep your ficus?
Answer Choices:
(a) refigerator
(b) screened porch
(c) clay pot
(d) shady places
(e) arboretum
 A: The answer must be a container that keeps the ficus in a shady place. Clay pots are used to keep plants in shady places. Therefore, the answer is clay pot (c).
 A: The answer must be a container in which ficus can be kept. Ficus are tropical plants. Therefore, the answer is clay pot (c).
 A: Clay pot is the only type of container listed. Therefore, the answer is clay pot (c).

Q: James was cooking s stew in his apartment kitchen.  Where might he look for a potato.
Answer Choices:
(a) restaurants
(b) casserole
(c) root cellar
(d) pockets
(e) pantry
 A: The answer must be a place where potatoes are stored. The answer is pantry (e).
 A: The answer must be a place where potatoes are stored. Pantries are the most likely place to store potatoes. Therefore, the answer is pantry (e).
 A: apartments do not often have root cellars, and are not restaurants. a pantry is closest to James inside his apartment. Therefore, the answer is pantry (e).

Q: If ships are in a storm and the sun comes out, what has happened?
Answer Choices:
(a) sinks
(b) cross river
(c) collide
(d) bottom out
(e) carry people
 A: The answer must be a thing that happens when ships are in a storm and the sun comes out. Ships are in a storm when they are in the middle of the ocean. Therefore, the answer is bottom out (d).
 A: The answer must be something that happens when the sun comes out. The sun is a source of energy. Therefore, the answer is bottom out (d).
 A: the sun comes out so its not a storm any more. Therefore, the answer is bottom out (d).

Q: Where do all marmot live before they are captured?
Answer Choices:
(a) switzerland
(b) spain
(c) vancouver island
(d) wilderness
(e) australia
 A: The answer must be a place where marmots live. Marmots live in the wilderness. Therefore, the answer is wilderness (d).
 A: The answer must be a place where marmots are found. The wilderness is the most likely place to find marmots. Therefore, the answer is wilderness (d).
 A: Marmots don't live in Australia or Switzerland. Therefore, the answer is wilderness (d).

Q: Where does one wait on a bench for something on tracks?
Answer Choices:
(a) train station
(b) state park
(c) crosswalk
(d) bus stop
(e) bus depot
 A: The answer must be a place where one waits for a train, bus, or other vehicle. The answer is train station (a).
 A: The answer must be a place where one can wait for a train. Train stations are places where trains stop. Therefore, the answer is train station (a).
 A: Only trains are on tracks. Therefore, the answer is train station (a).

Q: What can planning vacation lead to between a couple when they disagree?
Answer Choices:
(a) relaxation
(b) enjoying
(c) arguments
(d) going abroad
(e) spending money
 A: The answer must be something that can happen when a couple disagrees. Arguments are a common result of disagreements. Therefore, the answer is arguments (c).
 A: The answer must be a result of planning vacation. Arguments are a common result of planning vacation. Therefore, the answer is arguments (c).
 A: Arguments are the direct result of disagreements. Therefore, the answer is arguments (c).

Q: A sloth climbs slow, but it can eventually reach all the way up where?
Answer Choices:
(a) forest canopy
(b) universe
(c) tropical rainforest
(d) large trees
(e) nature preserve
 A: The answer must be a place where a sloth can climb. The forest canopy is the highest point in the forest. Therefore, the answer is forest canopy (a).
 A: The answer must be a place where a sloth can climb. The canopy of a forest is the highest point in a forest. Therefore, the answer is canopy of a forest (a).
 A: The sloth is a mammal that lives in the trees of South America. They have a slow and peculiar way of moving about. In fact, they are the slowest land mammals and travel very sluggishly on the ground (0.025 m/s-0.050 m/s). Therefore, the answer is canopy of a forest (a).

Q: Where in America is known for theaters?
Answer Choices:
(a) new york city
(b) shopping mall
(c) downtown area
(d) skating rink
(e) military base

 A: The answer must be a place where people go to see movies. Theaters are places where people go to see movies. Therefore, the answer is theater (a).
 A: The answer must be a place with a lot of theaters. New York City is known for its many theaters. Therefore, the answer is New York City (a).
 A: New York City is the main region in US known for theaters. Therefore, the answer is New York City (a).

Q: What does hearing someone play violin beautifully make you?
Answer Choices:
(a) buy earplugs
(b) inspiring
(c) guitar
(d) make music
(e) like music
 A: The answer must be something that makes you feel inspired. Hearing someone play violin beautifully makes you feel inspired. Therefore, the answer is inspiring (b).
 A: The answer must be a feeling which would cause someone to want to hear someone play violin beautifully. Music is a form of art. Therefore, the answer is inspiring (b).
 A: Hearing someone play a violin beautifully makes me inspired. Therefore, the answer is inspiring (b).

Q: What emotion does committing murder induce?
Answer Choices:
(a) fear
(b) go to jail
(c) problems
(d) cruel
(e) dead
 A: The answer must be a feeling that would cause someone to commit murder. Murder is a crime. Therefore, the answer is fear (a).
 A: The answer must be a feeling which would cause someone to commit murder. If someone feels fear, they are likely to want to avoid committing murder. Therefore, the answer is fear (a).
 A: The only feeling. Therefore, the answer is fear (a).

Q: Sitting to close while watching TV can cause what sort of pain?
Answer Choices:
(a) brain problems
(b) laziness
(c) get fat
(d) headache
(e) laughter
 A: The answer must be a pain that comes from sitting too close to the TV. Sitting too close to the TV can cause headaches. Therefore, the answer is headache (d).
 A: The answer must be a type of pain which can occur if you are sitting too close to the TV. Headaches are a common side effect of sitting too close to the TV. Therefore, the answer is headache (d).
 A: the process of aging in humans. Therefore, the answer is headache (d).

Q: What might people do to protect their legs from getting dirty on the farm?
Answer Choices:
(a) wear jeans
(b) milk cow
(c) believe in god
(d) avoid mud
(e) plant flowers
 A: The answer must be something that people do to protect their legs from getting dirty. Of the above choices, wearing jeans is the most logical. Therefore, the answer is jeans (a).
 A: The answer must be something that would protect legs from getting dirty on the farm. Jeans are used to protect legs from getting dirty. Therefore, the answer is wear jeans (a).
 A: Jeans cover over legs. Therefore, the answer is wear jeans (a).

Q: What types of stoves are there?
Answer Choices:
(a) these twos are irrelevent to the questions
(b) warm pot of beans
(c) brown meat
(d) warm room
(e) gas or electric
 A: The answer must be a type of stove. The stove is used to heat food. Therefore, the answer is gas or electric (e).
 A: The answer must be a type of stove. Stoves are used to warm rooms. Therefore, the answer is electric (e).
 A: stoves are gas or electric. Therefore, the answer is gas or electric (e).

Q: In what geological feature will you find fungus growing?
Answer Choices:
(a) shower stall
(b) toenails
(c) basement
(d) forrest
(e) cave
 A: The answer must be a place where fungus grows. The answer is cave (e).
 A: The answer must be a place where fungus grows. Fungus grows in caves. Therefore, the answer is cave (e).
 A: Rivers flow trough valleys. Therefore, the answer is cave (e).

Q: What is our sun considered to be the center of?
Answer Choices:
(a) universe
(b) solar system
(c) outer space
(d) star
(e) beans
 A: The answer must be something that is considered to be the center of the universe. The sun is the center of the solar system. Therefore, the answer is solar system (b).
 A: The answer must be a place where the sun is the center of. The sun is the center of the solar system. Therefore, the answer is solar system (b).
 A: It is the best option because it is the closest. Therefore, the answer is solar system (b).

Q: What are people in a library likely doing?
Answer Choices:
(a) talk to each other
(b) board ships
(c) study books
(d) suffer hunger
(e) playing games
 A: The answer must be something that people in a library are likely to be doing. People in a library are likely to be studying books. Therefore, the answer is study books (c).
 A: The answer must be something that is likely to happen in a library. Libraries are places where people study books. Therefore, the answer is studying books (c).
 A: Books shelved as study-skills: How to Become a Straight-A Student by Cal Newport, The Study Skills Handbook by Stella Cottrell, Make It Stick: The Scienc... Therefore, the answer is study books (c).

Q: A shark isn't something you have to worry about at the lake in the woods, but there are plenty of dangers in the what?
Answer Choices:
(a) great oceans
(b) great outdoors
(c) tropical waters
(d) pacific ocean
(e) gulf of mexico
 A: The answer must be something that is dangerous in the great outdoors. The great outdoors is a place where there are many dangers. Therefore, the answer is great outdoors (b).
 A: The answer must be a place where sharks are a danger. Lakes in the woods are not a place where sharks are a danger. Therefore, the answer is great outdoors (b).
 A: The great outdoors is the only choice where you would find a lake in the woods. Therefore, the answer is great outdoors (b).

Q: You can share files with someone if you have a connection to a what?
Answer Choices:
(a) freeway
(b) radio
(c) wires
(d) computer network
(e) electrical circuit
 A: The answer must be a way to share files. The answer is computer network (d).
 A: The answer must be a connection which allows you to share files. A computer network is a collection of computers which are connected to each other. Therefore, the answer is computer network (d).
 A: Files can be transmitted over a computer network, and not a freeway.  "wires" doesnt make any sense. Therefore, the answer is computer network (d).

Q: What can an injury happen to?
Answer Choices:
(a) emergency room
(b) doctor's office
(c) living thing
(d) bones
(e) accident
 A: The answer must be something that can happen to a living thing. The answer is living thing (c).
 A: The answer must be something that can happen to a living thing. If someone is injured, they are likely to want to go to the emergency room. Therefore, the answer is living thing (c).
 A: Living things feel pain, and therefore can be injured. Therefore, the answer is living thing (c).
\end{lstlisting}
\end{multicols}
}

\newpage
\section{Example Rationalizations on CQA}
We include a randomly sampled set of rationalizations which the model is able to produce before fine-tuning. We observe that sometimes, the model constructs an argument roughly of the form ``the answer must have \textit{a set of properties}. \textit{correct answer} has \textit{those properties}. therefore, the answer is \textit{correct answer}.'' This structure of argument is fairly standard, but given that the model originally answered those questions incorrectly, it resembles template-matching more than reasoning. The technique of rephrasing the question in a simpler way before answering it directly is also often effective. In general, the rationales from the pretrained model's rationalization appears to be of similar quality to the pretrained model's rationale generation. We reference the pretrained model here rather than the STaR-trained model, as the final iteration's rationales are the ones which the model struggled to learn from -- that is, if the rationales were good, then we might expect that the model would have already incorporated them into its rationale generation.

\begin{lstlisting}
Q: Why do I eat just enough when eating breakfast?
Answer Choices:
(a) gas
(b) full stomach
(c) feeling satisfied (CORRECT)
(d) have energy
(e) bloated
A: The answer must be a reason for eating just enough when eating breakfast. Eating just enough is a way to feel satisfied. Therefore, the answer is feeling satisfied (c).

Q: How is a dog likely to communicate with another dog?
Answer Choices:
(a) reproduce
(b) bark (CORRECT)
(c) do many things
(d) clone
(e) jump up
A: The answer must be a way for a dog to communicate with another dog. Dogs bark to communicate with each other. Therefore, the answer is bark (b).

Q: If a person is trying to weasel out of punishment for a crime, where are they likely to be?
Answer Choices:
(a) court room (CORRECT)
(b) cherry tree
(c) chicken coop
(d) natural history museum
(e) jail
A: The answer must be a place where a person would be trying to get out of punishment for a crime. The answer is court room (a).

Q: What would easily hide a snake?
Answer Choices:
(a) living room of a house
(b) kew gardens
(c) terrarium
(d) thick forest (CORRECT)
(e) tropical forest
A: The answer must be something that would hide a snake. The answer is thick forest (d).

Q: The person didn't like the varying size of the cake slices handed out, she demand more what?
Answer Choices:
(a) fairness (CORRECT)
(b) candies
(c) compliments
(d) well fed
(e) arguements
A: The answer must be a reason for the person to demand more cake. The person is not happy with the size of the cake slices. Therefore, the answer is fairness (a).

Q: In America, the courthouse of any given town has a likely location, where is it?
Answer Choices:
(a) electrical circuit
(b) getting married
(c) capital city
(d) center of town (CORRECT)
(e) michigan
A: The answer must be a location where a courthouse is located. The capital city is the location of the capital of a state. Therefore, the answer is center of town (d).

Q: Where might I find a bowl nestled among other bowls in the kitchen?
Answer Choices:
(a) refrigerator
(b) kitchen top
(c) cupboard (CORRECT)
(d) dishwasher
(e) apartment
A: The answer must be a place where a bowl is found. The kitchen is a place where bowls are found. Therefore, the answer is cupboard (c).

Q: Which state has blue ridge mountains that are called Blue Mountain?
Answer Choices:
(a) appalachians
(b) virginia
(c) pennsylvania (CORRECT)
(d) north carolina
(e) montana
A: The answer must be a state with mountains. The answer is Pennsylvania (c).

Q: Where could you find many radio studio?
Answer Choices:
(a) radio station
(b) country
(c) clear area
(d) office building
(e) large city (CORRECT)
A: The answer must be a place where many radio studios are located. Radio studios are used to broadcast radio programs. Therefore, the answer is large city (e).

Q: Where would someone bring you a cup?
Answer Choices:
(a) apartment
(b) closet
(c) restaurant (CORRECT)
(d) table
(e) party
A: The answer must be a place where someone would bring you a cup. A restaurant is a place where people bring cups. Therefore, the answer is restaurant (c).
\end{lstlisting}

\newpage
\section{STaR Without Rationalization (Rationale Generation Bootstrapping)}

For convenience, we include the rationale generation bootstrapping algorithm alone here. However, it is simply Algorithm~\ref{algostar}, with the blue components removed.

\begin{algorithm}
	\caption{Rationale Generation Bootstrapping (STaR without rationalization)} 
    \hspace*{\algorithmicindent} \textbf{Input} $M$: a pretrained LLM; dataset $\mathcal{D} = \{(x_i, y_i)\}_{i = 1}^D$ (w/ few-shot prompts)
	\begin{algorithmic}[1]
	    \State ${M_0} \leftarrow$ ${M}$ {\color{Gray} \# Copy the original model}
		\For {$n$ \textbf{in} $1...N$} {\color{Gray} \# Outer loop}
		    \State $(\hat{r}_i, \hat{y}_i) \leftarrow M_{n - 1}(x_i)\quad \forall i \in [1, D]$ {\color{Gray} \# Perform rationale generation}
		    \State $\mathcal{D}_n \leftarrow \{ (x_i, \hat{r}_i, y_i) \mid i \in [1, D] \land \hat{y}_i = y_i \}$ {\color{Gray} \# Filter rationales using ground truth answers}
		    \State $M_n \leftarrow \text{train}(M, \mathcal{D}_n)$ {\color{Gray} \# Finetune the original model on the correct solutions - inner loop}
		\EndFor
	\end{algorithmic}
	\label{algobase}
\end{algorithm}

\section{Case Study}
\begin{figure}[h]
\hspace{-10px}
\begin{subfigure}{.5\textwidth}
  \centering
    {
    \small
\begin{verbatim}
Q: John wants to eat at an Indian 
restaurant in on top of the space 
needle, where would he have to go?
Answer Choices:
(a) reservation (b) curry 
(c) britsh town (d) seattle 
(e) large city
A: The answer must be a place with 
a space needle. The space needle is
a landmark in Seattle. Therefore, 
the answer is Seattle (d).

Q: I put a towel in a cardboard box for 
the kitten, I wanted it to have what?
Answer Choices:
(a) farmhouse (b) zoos (c) cold place
(d) floor (e) warm place
A: The answer must be a place where the 
kitten would be comfortable. The kitten 
would be comfortable in a warm place. 
Therefore, the answer is warm place (e).

Q: Needlepoint requires many 
accessories, what holds the work taut?
Answer Choices:
(a) sewing (b) collection of folk art
(c) livingroom (d) frame (e) bathroom
A: The answer must be something that 
holds the work taut. The work is taut 
because it is held in place. Therefore,
the answer is frame (d).
\end{verbatim}
    }
  \caption{Problems where the model was able to solve questions which it had previously answered incorrectly.}
  \label{fig:sub-first}
\end{subfigure}
\hspace{10px}
\begin{subfigure}{.5\textwidth}
  \centering
    {
    \small
    \begin{Verbatim}[commandchars=+\[\]]
Q: They prided themselves on being a 
wealth  of knowledge, and that's why many 
chose to attend their what?
Answer Choices:
(a) book store (b) university (c) meeting
(d) class (e) encyclopedia
+sout[A: The answer must be a place where people]
+sout[go to learn about things. The answer is]
+sout[university (b).]
A: The answer must be a place where people
go to learn about things. Universities are
places where people go to learn about 
things. Therefore, the answer is 
university (b).

Q: You want a real estate agent that is 
straight forward with you, the last 
thing you want is for them to what a 
house's quality?
Answer Choices:
(a) cover up (b) offer price
(c) misrepresent (d) lie (e) falsify
+sout[A: The answer must be a way to determine]
+sout[the quality of a house. The answer is]
+sout[misrepresent (c).]
A: The answer must be something that 
would be a bad thing for the real estate
agent to do. The real estate agent would
be misrepresenting the house's quality
if they were to lie about it. Therefore,
the answer is misrepresent (c).
\end{Verbatim}
}
  \caption{Problems where the model was already able to answer them with few-shot reasoning, but was able to answer better after more iterations.}
  \label{fig:sub-second}
\end{subfigure}
\caption{Case study on STaR. We highlight some problems which the model is not initially able to solve, as well as rationales which are initially correct but improve. We note that for CQA, one is tasked with selecting the ``best'' answer, where multiple answers may be technically correct.} 
\label{fig:exampleresult}
\vspace{-5px}
\end{figure}

\newpage
\section{Bias and Faithfulness}
\label{biasexplanation}
It is important to note that STaR is designed to amplify the reasoning that leads to correct solutions on a given dataset. The implication of this with a dataset like CommonsenseQA is that if biases are ``useful'' in solving the dataset then they will be amplified. This is made worse by rationalization, as biased answers that the model may not naturally arrive at are, in a sense, pulled out of the model. The exact interaction between the bias in a dataset and the pre-existing bias learned by the model is unclear, and something to consider before real-world deployment of all large language models in general, but STaR-trained models in particular. 

We find some encouraging initial results on this however: for questions where gender is not relevant, the model appears to disregard it in its explanation. For example: ``Q: Where is a good place for a woman to store her sunglasses? $\rightarrow$ A: The answer must be a place where sunglasses are stored. Sunglasses are stored in purses. Therefore, the answer is purse (e).'' or ``Q: The only baggage the woman checked was a drawstring bag, where was she heading with it? $\rightarrow$ A: The answer must be a place where a drawstring bag is checked. The answer is airport (e).'' We believe this question warrants a much more comprehensive study.

One important challenge with models which seek to be interpretable or provide explanations for their reasoning is that of faithfulness. While our primary emphasis is not on the explainability benefits that STaR may bring, there is a fundamental challenge around evaluating explanations and rationales: namely, faithfulness \citep{herman2017promise,jacovi2020towards}. \citeauthor{jacovi2020towards} describe faithful explanations as those which ``accurately [represent] the
reasoning process behind the model’s prediction.'' While STaR encourages the use of reasoning in rationales which leads the model to correct answers, it is difficult, if not impossible, to ensure that the rationales reflect the model's internal processing. For example, it is straightforward to imagine the model implicitly selecting a particular answer immediately and then generating a rationale to justify that selected answer. This would allow a model to generate unbiased rationales while selecting answers in a biased way.

The fact that our model outperforms one fine-tuned to directly predict the answers, and ablation studies from papers such as \citet{wei2022chain} make it clear that the generation of a rationale before producing an answer non-trivially improves the model's answer quality. However, it is difficult to evaluate the degree to which any particular answer's rationale is faithful. However, we note that there problems are not unique to STaR, but are symptomatic of the difficulty of understanding large language models and in particular the rationales generated by large language models.

\section{Hyperparameters}
\label{hyperparameters}
GPT-J is a 28-layer decoder-only transformer, with an embedding size of 1024, 16 attention heads of dimension 256, and an FFN hidden layer of size 16384. It was pre-trained on the Pile~\citep{gao2020pile}, with a vocabulary size of 50.4K

In general, unless otherwise stated, we use a batch size of 8 sequences, each of length 1024. We also use packing, namely, packing the shorter examples to form longer sequences (up to length 1024) to improve TPU utilization. We do not use weight decay, and we train and sample on a single TPU-v3 node. We performed a hyperparameter search over learning rates from $10^{-7}$ to $10^{-4}$ using the Adam optimizer \cite{adam}. We found that $10^{-6}$ was consistently the best-performing learning rate. 

\newpage
\section{GSM8K Few-shot Prompt}
\label{gsm8kprompts}
We include the following few-shot prompts for GSM8K, based on the examples in \citet{cobbe2021training}.

\begin{lstlisting}
Q: Natalia sold clips to 48 of her friends in April, and then she sold half as many clips in May. How many clips did Natalia sell altogether in April and May?
A: Natalia sold 48/2 = <<48/2=24>>24 clips in May.
Natalia sold 48+24 = <<48+24=72>>72 clips altogether in April and May.
#### 72

Q: Betty is saving money for a new wallet which costs $100. Betty has only half of the money she needs. Her parents decided to give her $15 for that purpose, and her grandparents twice as much as her parents. How much more money does Betty need to buy the wallet?
A: In the beginning, Betty has only 100 / 2 = $<<100/2=50>>50.
Betty's grandparents gave her 15 * 2 = $<<15*2=30>>30.
This means, Betty needs 100 - 50 - 30 - 15 = $<<100-50-30-15=5>>5 more.
#### 5

Q: Julie is reading a 120-page book. Yesterday, she was able to read 12 pages and today, she read twice as many pages as yesterday. If she wants to read half of the remaining pages tomorrow, how many pages should she read?
A: Maila read 12 x 2 = <<12*2=24>>24 pages today.
So she was able to read a total of 12 + 24 = <<12+24=36>>36 pages since yesterday.
There are 120 - 36 = <<120-36=84>>84 pages left to be read.
Since she wants to read half of the remaining pages tomorrow, then she should read 84/2 = <<84/2=42>>42 pages.
#### 42

Q: Mark has a garden with flowers. He planted plants of three different colors in it. Ten of them are yellow, and there are 80% more of those in purple. There are only 25% as many green flowers as there are yellow and purple flowers. How many flowers does Mark have in his garden?
A: There are 80/100 * 10 = <<80/100*10=8>>8 more purple flowers than yellow flowers.
So in Mark's garden, there are 10 + 8 = <<10+8=18>>18 purple flowers.
Purple and yellow flowers sum up to 10 + 18 = <<10+18=28>>28 flowers.
That means in Mark's garden there are 25/100 * 28 = <<25/100*28=7>>7 green flowers.
So in total Mark has 28 + 7 = <<28+7=35>>35 plants in his garden.
#### 35

Q: Alexis is applying for a new job and bought a new set of business clothes to wear to the interview. She went to a department store with a budget of $200 and spent $30 on a button-up shirt, $46 on suit pants, $38 on a suit coat, $11 on socks, and $18 on a belt. She also purchased a pair of shoes, but lost the receipt for them. She has $16 left from her budget. How much did Alexis pay for the shoes?
A: Let S be the amount Alexis paid for the shoes.
She spent S + 30 + 46 + 38 + 11 + 18 = S + <<+30+46+38+11+18=143>>143.
She used all but $16 of her budget, so S + 143 = 200 - 16 = 184.
Thus, Alexis paid S = 184 - 143 = $<<184-143=41>>41 for the shoes.
#### 41

Q: Tina makes $18.00 an hour.  If she works more than 8 hours per shift, she is eligible for overtime, which is paid by your hourly wage + 1/2 your hourly wage.  If she works 10 hours every day for 5 days, how much money does she make?
A: She works 8 hours a day for $18 per hour so she makes 8*18 = $<<8*18=144.00>>144.00 per 8-hour shift
She works 10 hours a day and anything over 8 hours is eligible for overtime, so she gets 10-8 = <<10-8=2>>2 hours of overtime
Overtime is calculated as time and a half so and she makes $18/hour so her overtime pay is 18*.5 = $<<18*.5=9.00>>9.00
Her overtime pay is 18+9 = $<<18+9=27.00>>27.00
Her base pay is $144.00 per 8-hour shift and she works 5 days and makes 5 * $144 = $<<144*5=720.00>>720.00
Her overtime pay is $27.00 per hour and she works 2 hours of overtime per day and makes 27*2 = $<<27*2=54.00>>54.00 in overtime pay
2 hours of overtime pay for 5 days means she makes 54*5 = $270.00
In 5 days her base pay is $720.00 and she makes $270.00 in overtime pay so she makes $720 + $270 = $<<720+270=990.00>>990.00
#### 990
\end{lstlisting}

\newpage
\section{STaR GSM8K Solutions}
\label{gsm8ksolutions}
\begin{figure}[h]
\centering
\hspace{-20px}
\begin{subfigure}{\textwidth}
  \centering
    {
    \small
\hspace{3px}\includegraphics[width=\textwidth]{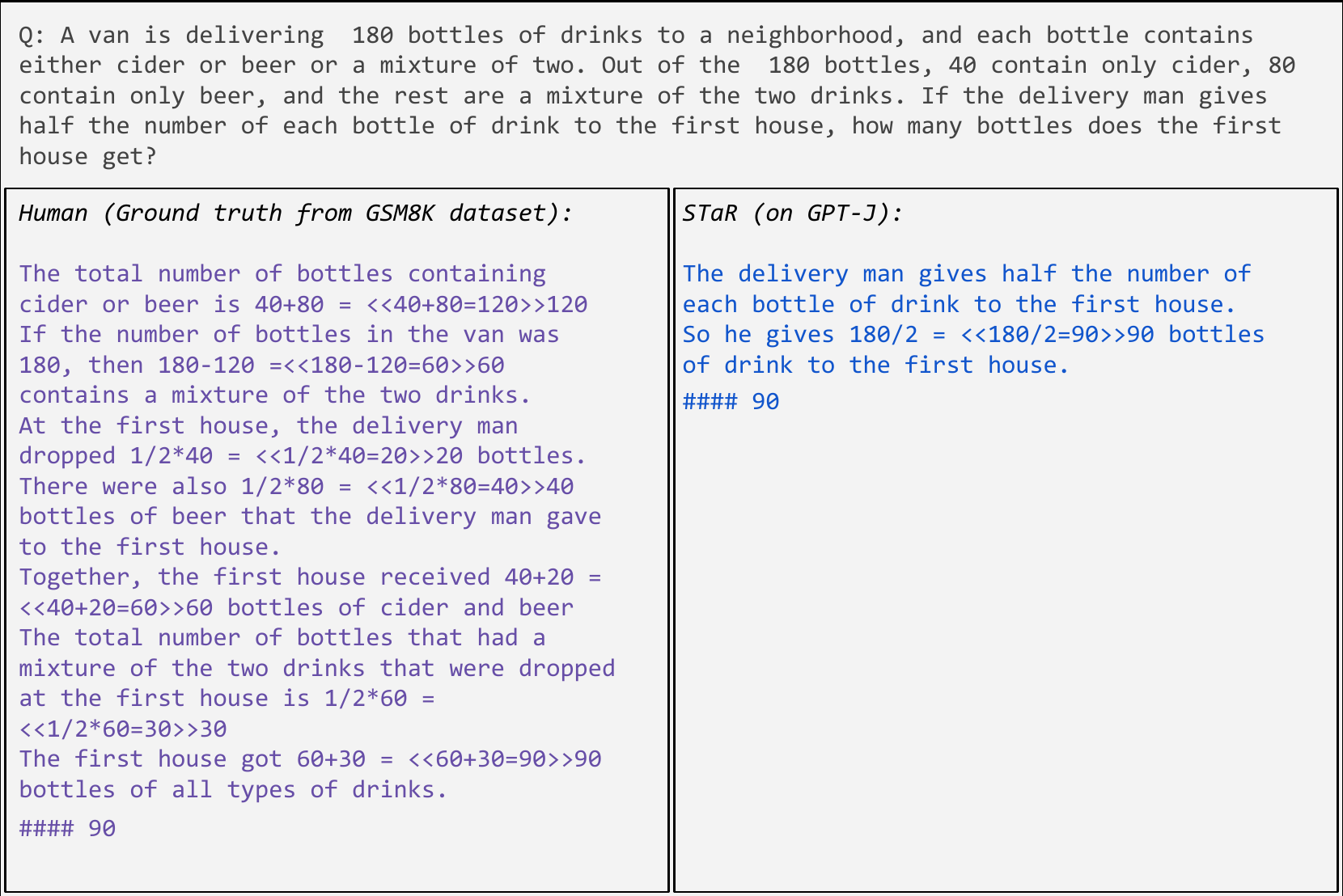}    }
\end{subfigure}
\caption{An example problem in the training set where STaR derives a significantly simpler solution than the ground truth.} 
\label{fig:shortcut}
\vspace{-10px}
\end{figure}

We observe some interesting patterns with the GSM8K solutions proposed by the STaR-trained model. Typically, when the solution takes substantially fewer calculation steps than the ground truth, it corresponds to an instance where the model accidentally answered the question correctly despite mistakes in its reasoning. In some cases, however, the model produces simpler solutions than those in the ground truth. One example is shown in Figure~\ref{fig:shortcut}.

\end{document}